\documentclass{article}

\usepackage[preprint]{neurips_2022}

\usepackage[utf8]{inputenc} 
\usepackage[T1]{fontenc}    
\usepackage[hidelinks]{hyperref}       
\usepackage{url}            
\usepackage{booktabs}       
\usepackage{amsfonts}       
\usepackage{nicefrac}       
\usepackage{microtype}      
\usepackage{xcolor}         
\usepackage{wrapfig}
\usepackage{sidecap}
\usepackage{xfrac}
\usepackage{float}
\usepackage{graphicx}
\usepackage{subcaption}
\usepackage{amsmath,amsfonts,bm}

\def\eqref#1{equation~\ref{#1}}

\def\1{\bm{1}}

\def\vu{{\bm{u}}}

\def\vx{{\bm{x}}}

\DeclareMathAlphabet{\mathsfit}{\encodingdefault}{\sfdefault}{m}{sl}
\SetMathAlphabet{\mathsfit}{bold}{\encodingdefault}{\sfdefault}{bx}{n}

\newcommand{\R}{\mathbb{R}}

\newcommand{\M}{\mathcal{M}}

\DeclareMathOperator{\Tr}{Tr}

\newtheorem{theorem}{Theorem}[section]
\newtheorem{proposition}[theorem]{Proposition}

\newtheorem{definition}[theorem]{Definition}

\title{Density estimation on smooth manifolds with normalizing flows}

\author{%
  Dimitris Kalatzis \thanks{Section for Cognitive Systems, Technical University of Denmark} \\
  \texttt{dika@dtu.dk} \\
   \And
   Johan Ziruo Ye \footnotemark[1]\\
   \texttt{ziruo@gmail.com} \\
   \And
   Alison Pouplin \footnotemark[1]\\
   \texttt{alpu@dtu.dk} \\
   \And
   Jesper Wohlert \\
   \texttt{jesper@wohlert.nu} \\
   \And
   S{\o}ren Hauberg \footnotemark[1]\\
   \texttt{sohau@dtu.dk} \\
}

\begin{document}

\maketitle

\begin{abstract}
    We present a framework for learning probability distributions on topologically non-trivial manifolds, utilizing normalizing flows. Current methods focus on manifolds that are homeomorphic to Euclidean space, enforce strong structural priors on the learned models or use operations that do not easily scale to high dimensions. In contrast, our method learns distributions on a data manifold by “gluing” together multiple local models, thus defining an open cover of the data manifold. We demonstrate the efficiency of our approach on synthetic data of known manifolds, as well as higher dimensional manifolds of unknown topology, where our method exhibits better sample efficiency and competitive or superior performance against baselines in a number of tasks.
\end{abstract}

\section{Introduction}
    Normalizing flows \citep{pmlr-v37-rezende15, DBLP:journals/jmlr/PapamakariosNRM21} provide an elegant framework for modelling complex, multimodal probability distributions. Normalizing flows comprise a base distribution $P_U$ on a latent space $U$ and a diffeomorphism, which provides a 1-to-1 mapping of points from the data space to the latent space according to this base distribution. Given a data point $\vx$, the marginal likelihood can be computed via the change of variables formula $p(\vx) = p(\vu) \lvert \det J_{f}(\vu) \rvert^{-1} = p(\vu) \lvert \det J_{f^{-1}}(\vx) \rvert$ with $\vx = f(\vu)$. Typically, the base distribution $P_U$ is a normal or a uniform distribution, both of which are defined in Euclidean space.
    
    Real world data, however, often lie on a manifold, with examples including protein structures \citep{hamelryck2006sampling, boomsma2008generative}, geological data \citep{peel2001fitting, karpatne2018machine} or graph-structured and hierarchical data \citep{steyvers2005large, roy2007learning}. Diffeomorphisms preserve the topological properties of their domain and therefore modelling the density of manifold-valued data is a known failure mode of flows, due to the topological mismatch between the target distribution $P_{X^{\star}}$ and the base distribution $P_U$ \citep{NEURIPS2019_21be9a4b, DinhRADApproach2019, cornish20a}. In response, recent works have constructed flows for specific manifolds, such as tori, spheres and hyperbolic spaces \citep{rezende20a, bose20a}. \looseness=-1
    
    Still, in many realistic situations one may not know the topological properties of a given data set a priori, but one may reasonably assume an underlying manifold structure. Such cases generally fall under the \emph{manifold hypothesis} \citep{fefferman2016testing}, an important heuristic in machine learning, which states that high dimensional data can be described by a low dimensional submanifold embedded in the observation space. \citet{brehmer2020flows} propose to learn the shape of the manifold via learning a (single) chart to it, however this implies that the manifold's topological structure is Euclidean. Another set of works \citep{lou2020neural, mathieu2020riemannian, falorsi2020neural, rozen2021moser} exploit local geometric information to learn distributions on embedded submanifolds with non-Euclidean topology but these operations do not easily scale to high dimensions. So the question then emerges: Can flow models learn a probability distribution on manifolds with complex topology and also scale to higher dimensions? 
    
\begin{SCfigure} 
        \centering
        \begin{subfigure}[b]{0.2\textwidth}
            \centering
            \footnotesize{Ground truth}\\\vspace{2mm}
            \includegraphics[width=\textwidth]{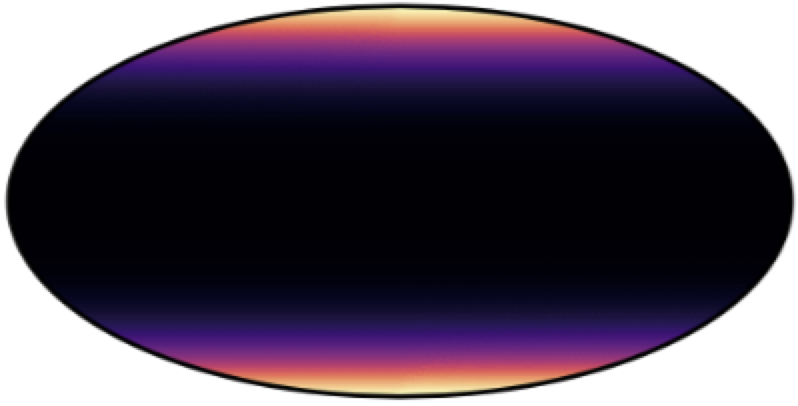}
        \end{subfigure}
        \begin{subfigure}[b]{0.2\textwidth}
            \centering
            \footnotesize{Multi-chart flows}\\\vspace{2mm}
            \includegraphics[width=\textwidth]{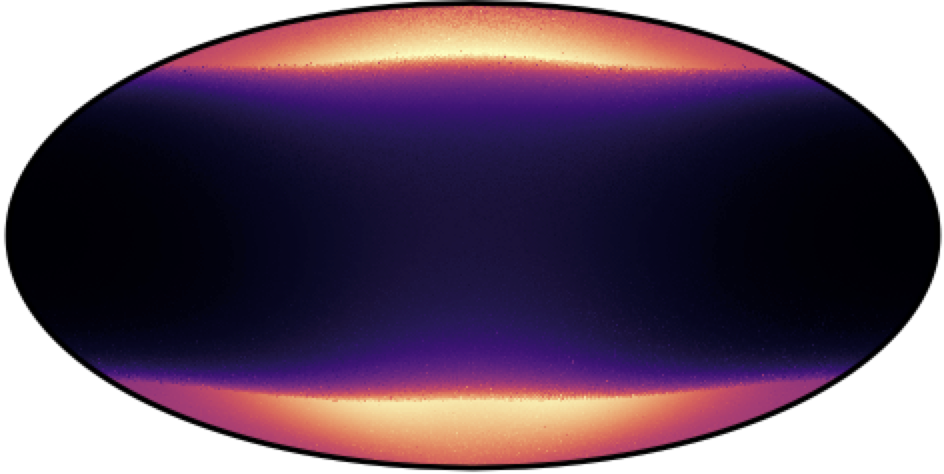}
        \end{subfigure}
        \begin{subfigure}[b]{0.2\textwidth}
            \centering
            \footnotesize{$\M$-flow}\\\vspace{2mm}
            \includegraphics[width=\textwidth]{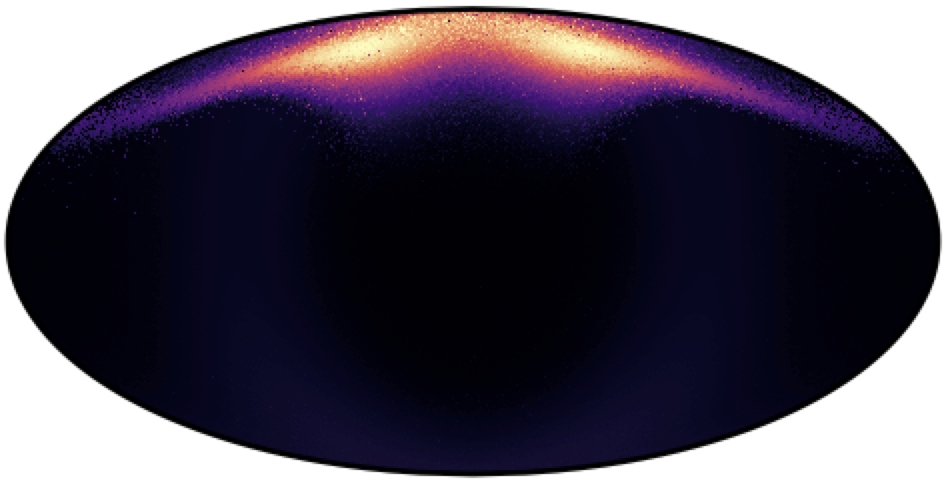}
        \end{subfigure}
        \caption{A bimodal distribution on a sphere. Contrary to our approach, single-charted models (like the $\M$-flow model \citep{brehmer2020flows}) struggles to push probability mass to cover both modes.}
        \label{fig:vmf-mix}
    \end{SCfigure}

    Our approach leverages the class of functions typically learned by flow models to learn a collection of smooth coordinate charts that cover the data manifold. Unlike existing methods, which do not make assumptions on manifold topology, we are able to learn probability distributions on data manifolds with complex (non-Euclidean) topological structure (Fig.~\ref{fig:vmf-mix}). Furthermore, in contrast to methods that depend on local geometry, our model scales to high dimensional non-Euclidean data. Finally, we are able to achieve competitive or superior performance in all tasks with better sample efficiency and faster runtimes than most of our baselines.

\section{Smooth manifolds}
    \label{sec:smooth-manifolds}
    To make subsequent exposition clearer we will briefly review a few basic notions. We begin with the definition of a smooth manifold, which is central to the construction of our model in the next section.
    
    \begin{definition} \label{def:topological_manifold}
        A smooth manifold $\M$ of dimension $d$ is a topological space that is locally Euclidean, i.e. each point of $\M$ has a neighborhood $U$ which is diffeomorphic to an open subset of $V \subset \R^d$.
    \end{definition}
    
    \begin{wrapfigure}[13]{r}{0.5\textwidth}
        \vspace{-6mm}
        \begin{center}
            \includegraphics[width=0.5\textwidth]{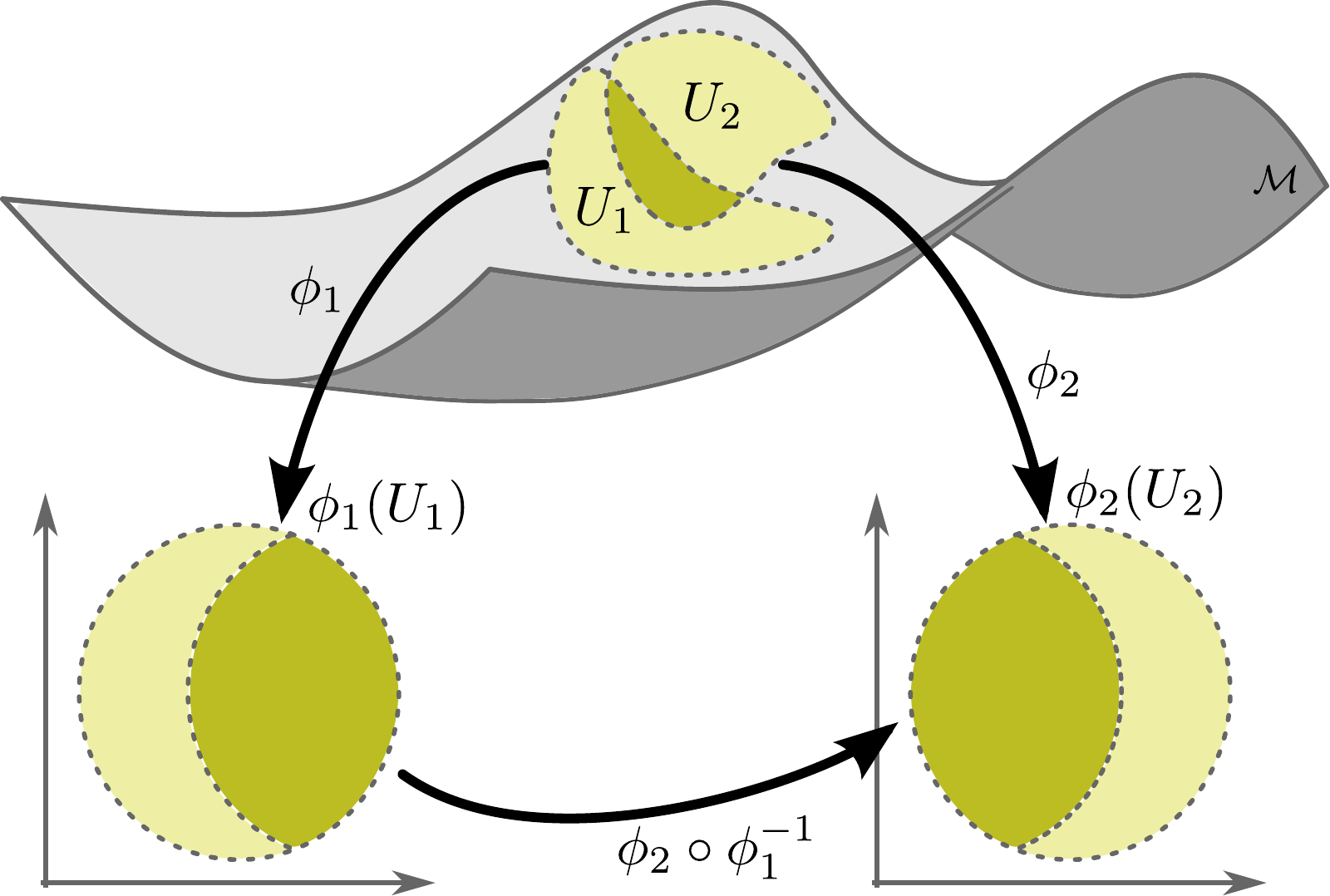}
        \end{center}
        \vspace{-1mm}
      \caption{Smoothly compatible charts.}
      \label{fig:chart-compatibility}
    \end{wrapfigure}
    
    We can now formalize the ``locally Euclidean'' property of a smooth manifold by introducing 
    \emph{smooth local coordinate charts} on $\M$.
    
    \begin{definition} \label{def:coordinate-charts}
        Given a $d$-dimensional topological manifold $\M$, a smooth coordinate chart on $\M$ is a pair $(U, \phi)$, where $\phi\colon U \rightarrow V$ is a diffeomorphism between the open subsets $U \subset \M$ and $V \subset \R^d$.
    \end{definition}
    
    To have local coordinates for every point on $\M$ we can define a collection of smooth coordinate charts that covers $\M$. This collection is called a \emph{smooth atlas}. This construction is necessary to define smooth functions (such as probability density functions) and perform gradient-based optimization on $\M$, since for any smooth coordinate chart $(U, \phi)$ and a function $f: \M \rightarrow \R$, the composition $f \circ \phi^{-1}: V \rightarrow \R$ is smooth. It further allows us to account for points occurring in overlapping charts without issues with regard to smoothness, since given two smooth coordinate charts $(U_1, \phi_1), (U_2, \phi_2)$ with $U_1 \cap U_2 \neq \emptyset$, the composition $\phi_2 \circ \phi_1^{-1}$ is smooth and invertible. These charts are then called \emph{smoothly compatible} (see Fig.~\ref{fig:chart-compatibility}).
    
    In this work we are considering a smooth manifold $\M$ of dimension $d$, embedded in some Euclidean space $\R^D$ with $d < D$. Embedded submanifolds can be defined as the images of \emph{smooth embeddings}. 
    
    \begin{definition}
        A smooth embedding is a smooth immersion (i.e. a map, with Jacobian that is full rank everywhere), which is also a diffeomorphism onto its image.
    \end{definition}
    
\begin{wrapfigure}[14]{r}{0.5\textwidth}
        \vspace{-15mm}
        \begin{center}
            \includegraphics[width=0.3\textwidth]{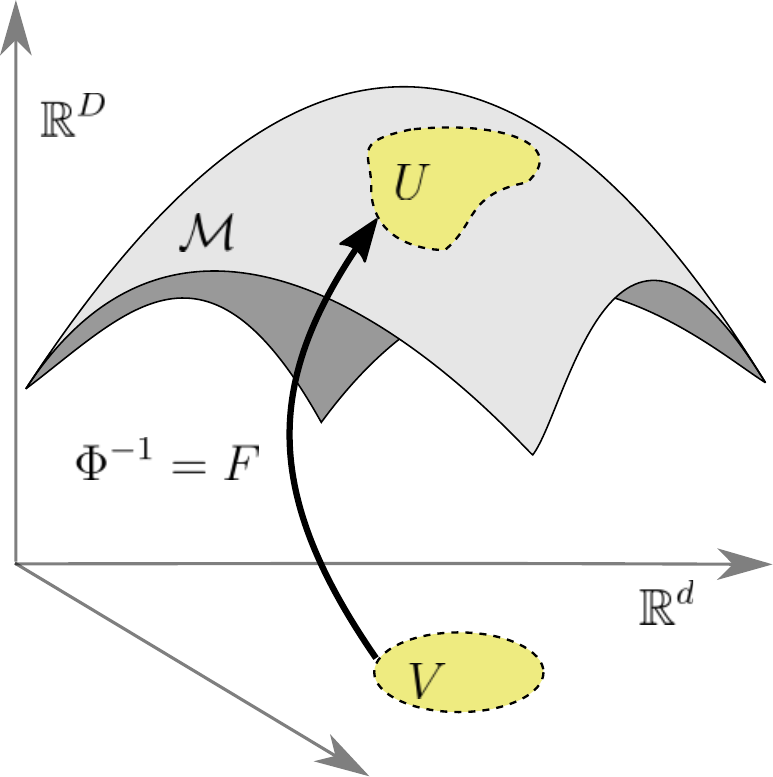}
        \end{center}
      \caption{A neighborhood $U$ of an embedded submanifold $\M \subset \R^D$ is the image of a smooth embedding $F: V \rightarrow \R^D$, with $V \subset \R^d$.}
      \label{fig:smooth-embedding}
    \end{wrapfigure}

    More specifically, a neighborhood $U \subset \M$ can be expressed as the image of a smooth embedding $F: V \rightarrow \R^D$, with $V \subset \R^d$ (see Fig.~\ref{fig:smooth-embedding}). Smooth embeddings are diffeomorphisms onto their image and as such, invertible when their codomain is restricted to it. Thus, the open subset $U \subset \M$ inherits the Euclidean topology of $V$ and we can define local coordinates on $U$, through the coordinate chart $(U, \phi)$ with $\phi = F^{-1}: \R^D \rightarrow V$ by restricting the domain of $F^{-1}$ to $U$.
    
\section{A multi-charted approach to density estimation on manifolds} \label{sec:mcf}
    We now present our main contribution, \emph{Multi-chart flows} (MCF). We introduce the construction of density functions on smooth manifolds and subsequently discuss training, inference and the generative process.

    \subsection{Model specification} \label{subsec:model}
    
    Given a local coordinate chart $(U, \phi)$ on the manifold, a probability density $p_U$ supported on a neighborhood $U \subset \M$ can be expressed through the change of variables formula:
    \begin{align}
        p_{U}(\vx) = p_{V}(\vu) |\det G(\vu)|^{-\frac{1}{2}}
        \label{eq:neighborhood-density}
    \end{align}
    where $p_{V}$ denotes a simple base density (e.g.\@a standard Gaussian) over the Euclidean subset $V$, $\vu = \phi(\vx)$ and $G = J_{\phi^{-1}}^{T} J_{\phi^{-1}}$ is induced by the smooth embedding $\phi^{-1}$ with the corresponding Jacobian matrix $J_{\phi^{-1}} \in \mathbb{R}^{D \times d}$. Here the more general form of the volume form is used, since $\phi^{-1}$ is injective. We seek to construct a probability density function $p_{\M}: \M \rightarrow \R$ over the manifold, by ``gluing" together multiple local models defined in subsets $U \subset \M$. To achieve this we will turn to a partition of unity construction \citep{strichartz2003guide, lee2013smooth} of such a density function. Let $\{U_i\}_{i=1}^K$ be an open cover of $\M$. Partitions of unity are families $\{f_i\}_{i=1}^{K}$, of continuous functions $f: \M \rightarrow \R$ with $\text{supp} f_i \subseteq U_i$ that satisfy the following:
    \begin{enumerate}
        \item In a neighborhood around a point $\vx \in \M$, only a finite subset of $\{f_i\}$ are non-zero.
        \item $\sum_{i=1}^K f_i(\vx) = 1$.
    \end{enumerate}
    
    As such, we can construct our density function $p_\M$ over the manifold by ``blending" together the density functions $p_U$ defined in local neighborhoods/coordinate patches on the manifold (eq.~\ref{eq:neighborhood-density}). Thus, with $i=1, \dots, K$ denoting the index of the neighborhood and $K$ the number of the overall neighborhoods in our cover of $\M$, which we treat as a hyperparameter we have:
    \begin{align}
        p_{\M}(\vx)
          &= \sum_{i=1}^K w_i p_{U_i}(\vx)
           = \sum_{i=1}^K w_i p_{V_i}(\vu)|\det G_i(\vu)|^{-\frac{1}{2}},
          \label{eq:manifold-density}
    \end{align}
    
    where $\sum_{i=1}^K w_i = 1$ and $\vu = \phi_i(\vx)$. We can furthermore normalize $w_i p_{U_i}(\vx)$ to satisfy the second condition of the partition of unity. This construction is convenient since it simultaneously allows us to define an open cover over our data manifold, which we can use as a smooth atlas, and removes the need to explicitly learn a reconstruction of the embedded manifold. The overall topological structure is preserved by constructing the manifold from locally Euclidean models. Furthermore, we avoid continuity/differentiability issues at the neighborhood boundaries. Because we are using flow models for our coordinate maps $\phi_i$, smooth chart compatibility is ensured by construction, since for overlapping coordinate charts  $(U_1, \phi_1), (U_2, \phi_2)$, the composition $\phi_2 \circ \phi_1^{-1}$ is a diffeomorphism as it is a composition of diffeomorphisms.
    
    \subsection{Introducing a lower bound to the density} \label{sec: lower-bound}
    While the determinant term in eq.~\ref{eq:manifold-density} can be computed exactly, it involves evaluating $G = J_{\phi_i^{-1}}^T J_{\phi_i^{-1}}$, which is prohibitively expensive even for a modest number of dimensions, since computing the determinant is an $O(d^3)$ operation. We introduce a lower bound to the log likelihood contribution of each chart (eq.~\ref{eq:neighborhood-density}), thereby lower bounding the complete data log likelihood (eq.~\ref{eq:manifold-density}). We will replace the determinant with the trace of $G$ which is an $O(d)$ operation. A sketch of a proof follows, with all details in Appendix~\ref{app:proof}. We drop neighborhood indices $i$ and for the log likelihood in a given coordinate patch $U$ with coordinate map $\phi$, we denote the singular values of $J_{\phi^{-1}}$ by $\{ s_i \}_{i=1}^d$ and we have:
    \begin{align}
      \log p_{U}(\vx)
        &= \log p_V(\vu) - \frac{1}{2}\log \det |G(\vu)| \\
        &= \log p_V(\vu) - \frac{1}{2} \sum_{i=1}^{d} \log s_i^2.
    \end{align}
    Using Jensen's inequality with uniform weights $a_i = 1/d$ we can bound this density by:
    \begin{align}
        \log p_U(\vx)
          &\geq \log p_V(\vu) - \frac{d}{2} \log \left (\sum_{i=1}^{d} s_i^2 \right ) + c \\
          &= \log p_V(\vu) - \frac{d}{2} \log \Tr[(J_{\phi^{-1}}(\vu))^{T} J_{\phi^{-1}}(\vu)] \nonumber \\ 
          &+ c,
    \end{align}
    where $c = d\log(d)/2$ is a constant.
    We can compute the trace efficiently using Hutchinson's estimator \citep{hutchinson1989stochastic}, arriving at:
    \begin{align}
        \log p_U(\vx) &\geq \log p_V(\vu) - \frac{d}{2} \log \mathbb{E}_{p(\boldsymbol{\epsilon})}\left[ ||J_{\phi^{-1}}^{T} \boldsymbol{\epsilon}||_2^2 \right]
    \end{align}
    with $\boldsymbol{\epsilon} \sim \mathcal{N}(0,I_D)$. Because for all $i$ we have $w_i \in [0, 1]$, inequality~\ref{eq:lower-bound-hutchinson} below holds for all neighborhoods $U_i$, and by extension the lower bound holds for the overall data log likelihood on the manifold:\looseness=-1
    \begin{align}
        \log p_{\M}(\vx) &= \log \sum_i^K w_i p_{U_i}(\vx) 
        = \log \sum_i^K w_i p_{V_i}(\vu) \det |G_i(\vu)|^{-\frac{1}{2}} \\
        &\geq \log \left[ C \cdot \sum_i^K w_i p_{V_i}(\vu) \mathbb{E}_{p(\boldsymbol{\epsilon})}\left[ ||J_{\phi^{-1}}^{T} \boldsymbol{\epsilon}||_2^2 \right]^{-\frac{d}{2}} \right]
        \qquad \text{ with } C = d^{d/2}.
        \label{eq:lower-bound-hutchinson}
    \end{align}
    
    \begin{figure}[t!]
        \begin{center}
            \includegraphics[width=0.65\textwidth]{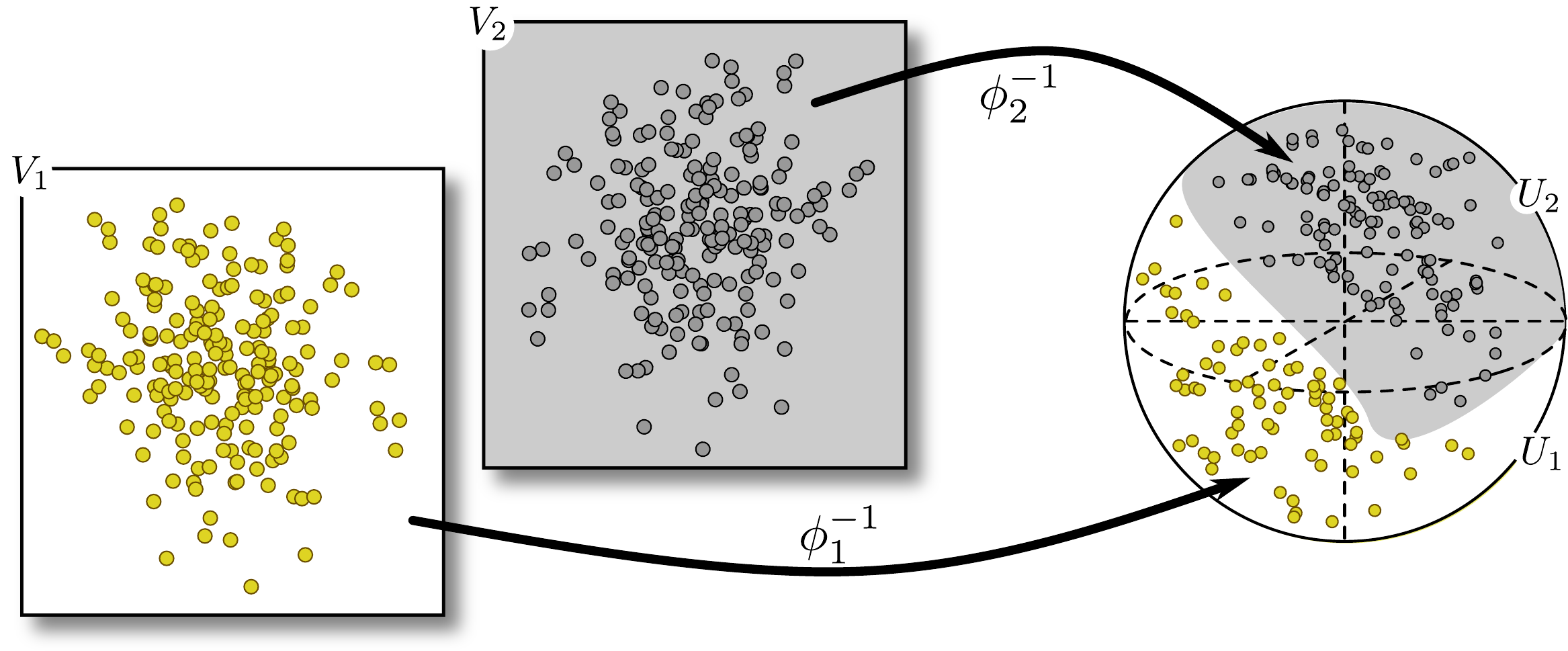}
        \end{center}
        \vspace{-3mm}
      \caption{Overview of the sampling/generative process proposed in Multi-chart flows. First, the index $k$ of some Euclidean subset $V_k$ is sampled. Then, points sampled in the lower dimensional Euclidean spaces $V_k$ are mapped onto the embedded data manifold by the inverse coordinate maps, $\phi_i^{-1}$.}
      \label{fig: sampling}
    \end{figure}

\subsection{Training}
    We train our model using maximum likelihood estimation on the lower bounded density (eq.~$\ref{eq:lower-bound-hutchinson}$). As mentioned before we do not need an explicit manifold learning/reconstruction step. We, furthermore, construct our coordinate maps as embeddings. To construct such a map using flow models, we append $D - d$ zeros to the base variable $\vu \in \R^d$ and map to $\M \subset \R^D$ with $\phi^{-1}$. We denote this ``augmented'' variable by $\vu' = [u_{1}, \dots, u_{d}, 0, \dots, 0]^{T} \in \R^D$ and for the remainder of the paper we will use this symbol to refer to this construction. As for the forward maps $\phi: U \subset \M \rightarrow V \subset \R^d$ we follow the strategy of \citet{beitler2021pie}, where we split the dimensions of the observed variable $\vx$ into the intrinsic manifold dimensions $d$ and the directions normal to the manifold $D - d$. We then use the map $\phi$ to map the manifold dimensions to the base distribution $p_V$ in the Euclidean subset $V$ and the orthogonal directions to a distribution $p_{V^{\perp}}$, which is tightly centered around $0$, e.g. a zero-centered Gaussian with $\sigma^2 = 0.01$. This way we can define the map $\phi$ as a projection from the manifold to the Euclidean domain $V_i$. Crucially and in contrast to \citet{beitler2021pie} we train on the more general form of the change of variables employing the correct volume measure induced by our embeddings $\phi^{-1}$, i.e. $dV = \sqrt{G(\vu)} d\vu$, with $G$ a Riemannian metric tensor defined as: $G = J_{\phi^{-1}}^{T} J_{\phi^{-1}}$. Although it is flexible, this construction deprives us of the ability to directly compare likelihoods across models, since the volume measure will depend on the chart parameterization of the manifold through the inverse coordinate map $\phi^{-1}$.

\subsection{Sampling}
    To determine the mixture probabilities for our model we use a neural network during training, in other words $\boldsymbol{w} = \text{NN}(\vx; \theta)$ with $\theta$ denoting the neural network parameters, $\boldsymbol{w}$ a normalized vector in $\R^K$ and $K$ the number of coordinate charts. To be able to sample from our model however, we assume a Categorical distribution over the charts and keep estimates of these probabilities throughout training, by simply normalizing the counts of coordinate chart assignments over the whole dataset. Denoting the index of a data point by $n = 1, \dots, N$, the index of a coordinate chart by $i = 1, \dots, K$ and the number of data points assigned to coordinate chart $i$ by $N_i$:
    \begin{align}
        \tilde{w}_i &= \frac{N_i}{N} \\
        p(\boldsymbol{c}; \boldsymbol{\tilde{w}}) &= \prod_{i=1}^K \tilde{w}_i^{\boldsymbol{c}_i}
    \end{align}
    Then, we can sample from the model through ancestral sampling, where we first sample chart $\boldsymbol{c}_k \sim p(\boldsymbol{c}; \boldsymbol{\tilde{w}})$, then the $d$-dimensional latent variable $\vu$, append $D-d$ zeros to get $\vu'$ and map to $\M \subset \R^D$ with $\phi_k^{-1}$ (Fig.~\ref{fig: sampling}). 
    
\section{Related work}
    \paragraph{Learning the manifold structure.}
    \citet{brehmer2020flows} propose learning the topological structure of the manifold separately from learning the probability distribution on it and so they split training into two distinct phases. Initially they learn a reconstruction of the data manifold via an embedding $g: \M \rightarrow \R^{D}$, which can be considered a composition $f \circ \phi$ with $\phi: \M \rightarrow \R^{d}$ the manifold chart and $f: \R^{d} \rightarrow \R^{D}$ a smooth, injective map. Then they learn the density on the manifold via a transformation $h: \R^d \rightarrow \R^d$. A crucial limitation is that the data manifold is assumed to be covered by the single chart $\phi$, i.e.\ it is homeomorphic to Euclidean space. The model, thus, cannot represent non-trivial manifolds. \citet{lou2020neural} proposed another closely related method treating the exponential map as a chart. Since the exponential map $\exp: T_{x}\M \rightarrow \M$ is a local diffeomorphism between the (Euclidean) tangent space at $x$ and the manifold $\M$, it is treated as a chart $\phi$ centered at $x$. They learn a vector field in $T_{x}\M$ by solving a local ODE for a short time interval, which is mapped onto $\M$ by the expmap. They use the inverse chart, $\log: \M \rightarrow T_{x'}\M$, to map to the new tangent space centered at $x'$ and repeat the process. In principle, this scheme is general,  but in practice, the exponential and logarithmic maps are prohibitively expensive for high dimensional manifolds. \citet{rozen2021moser} propose Moser flow (MF), a generative model where the learned density consists of a source distribution minus the divergence of a neural network. Therefore, their suggested model falls within the broader family of continuous normalizing flows (CNFs), however they approximate the local divergence operator instead of solving the ODE, achieving significant speedups against CNF-based models (such as e.g. \citet{grathwohl2018ffjord, mathieu2020riemannian}) in low dimensions. An important limitation, however, is that the divergence is computationally expensive to approximate in high dimensions, limiting the applicability of MF to general high dimensional settings. Finally, works that learn an atlas of the manifold have appeared in the literature. \citet{DBLP:journals/tip/NascimentoSML14} use Gaussian processes for the chart maps which are combined probabilistically to form an atlas, \citet{DBLP:conf/cvpr/PitelisRA13} combine local linear models into an atlas by minimizing a regularized reconstruction error that encourages a small number of charts. \citet{DBLP:conf/nips/Brand02} uses a mixture of kernel-based linear projections to build a common coordinate system of connected Euclidean patches. Finally, \citet{schonsheck2019chart} propose autoencoder-based coordinate maps to construct their atlas.
    
    \paragraph{Flows on fixed manifolds.}
    A related body of work pertains to flows on manifolds with a priori known topological structure. \citet{rezende20a} construct flows defined on circles, tori and spheres through projective transformations, as well as by adapting Euclidean models, such as autoregressive flows \cite{papamakarios2017masked} and spline flows \cite{muller19, durkan19}. Flow-based models defined in hyperbolic space were presented by \citet{bose20a}, wherein two variants are proposed, which use parallel transport of vectors and repeated calls to the exponential and logarithmic maps to map between the tangent bundle and the manifold. A more general method of learning a flow on a manifold was proposed by \citet{gemici2016normalizing}, which assumes knowledge of a coordinate chart $\phi: \M \rightarrow \R^d$ and an embedding $g: \R^d \rightarrow \R^D$ with $d < D$. A limitation here is that $\M$ needs to be homeomorphic to $\R^d$, since it is described by a single chart. When $\phi$ and $g$ are learned we arrive at the models presented by \citet{brehmer2020flows}. We generalize this setting by learning transformations between patches of the manifold $\M$ and subsets of Euclidean space $\R^D$.
    
    \begin{figure}[t]
        \centering
        \begin{tabular}{cccc}
        Target & MCF (ours) & NMODE & NCPS \\
        \toprule
        \includegraphics[width=0.22\textwidth]{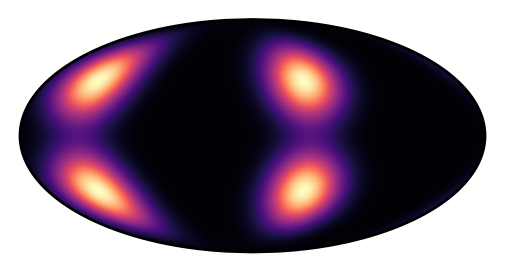} &
        \includegraphics[width=0.22\textwidth]{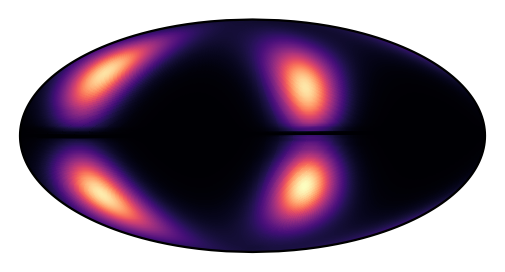} &  
        \includegraphics[width=0.22\textwidth]{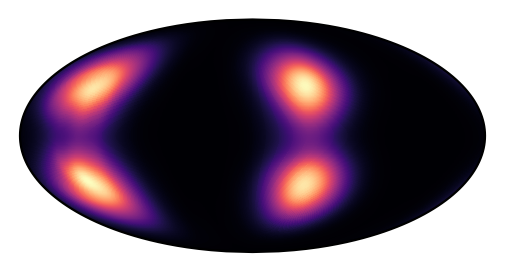} &
        \includegraphics[width=0.22\textwidth]{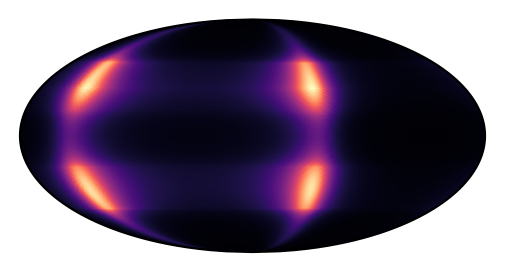} \\
        \includegraphics[width=0.22\textwidth]{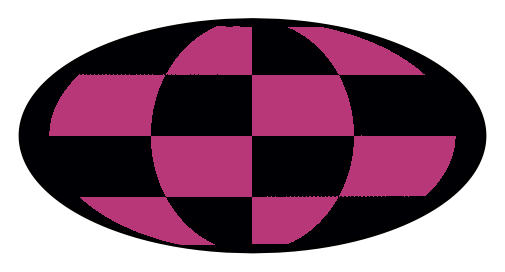} & \includegraphics[width=0.22\textwidth]{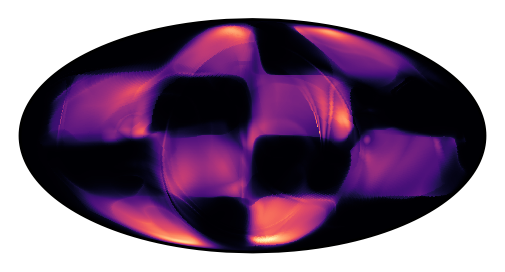} & 
        \includegraphics[width=0.22\textwidth]{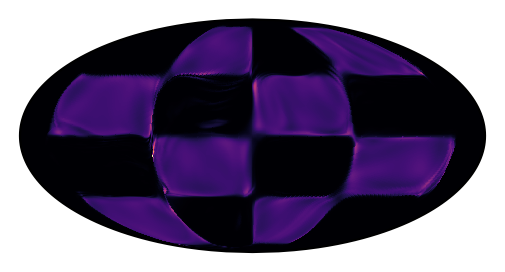} &
        \includegraphics[width=0.22\textwidth]{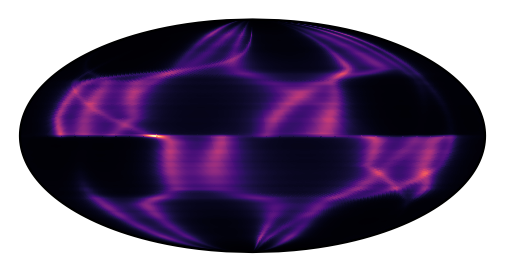}
        \end{tabular}
        \caption{Density estimation on the sphere $\mathbb{S}^2$ visualized via the Mollweide projection. Baselines:
        Neural Manifold ODEs (NMODE) by \citet{lou2020neural} and NCPS by \citet{rezende20a}}
        \label{fig:synthetic-sphere}
    \end{figure}
    
\section{Experiments} \label{sec:experiments}
    
    \subsection{Qualitative experiments: Estimation of synthetic densities on 2D manifolds} \label{sec:synthetic}
    For our first experiment we trained our model, denoted MCF (Multi-chart flows), on synthetic densities on the sphere $\mathbb{S}^2$, a 2D manifold with well studied topological structure. Our baselines were chosen among models that encode topological information as structural priors by way of a prescribed chart to access the manifold, and models that generally rely on the exponential map which still encodes local topological information on the manifold. Of the former, we chose the recursive circular spline flow (NCPS) \cite{rezende20a}. As for the latter, we chose neural manifold ODEs (NMODE) \cite{lou2020neural}. Results can be seen in Figure~\ref{fig:synthetic-sphere}. Our approach achieves improved performance over NCPS and performs on par with NMODE at significantly reduced running times (see section~\ref{sec:runtimes}). For the ``four wrapped normals" dataset (Fig.~\ref{fig:synthetic-sphere} top row) MCF uses two coordinate charts and each coordinate map comprises two rational quadratic (RQ) coupling layers interspersed with LU-decomposed, invertible linear maps. For the ``checkerboard" dataset (Fig.~\ref{fig:synthetic-sphere} bottom row), MCF uses four coordinate charts, with each coordinate map comprising three RQ coupling layers interspersed with LU-decomposed linear maps. Complete experimental details can be found in Appendix~\ref{app:synthetic}
    
    \subsection{Qualitative experiments: Estimation of real world densities on 2D manifolds}
    
\begin{figure}[t!]
        \centering
        \includegraphics[width=0.85\textwidth]{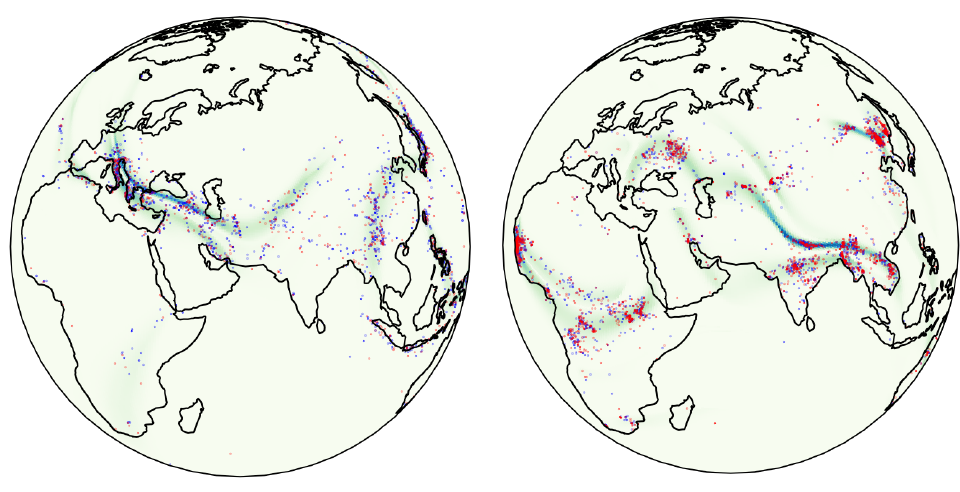}
        \caption{Density estimation results on real world densities. Learned density of the \emph{earthquakes} (left) and \emph{fires} (right) geological data as distributions on a sphere. Blue points are sampled from the training set, while red points are sampled from the evaluation set.}
        \label{fig:geo}
    \end{figure}

    We next examine a scenario of real world densities. Our datasets contain the locations of two types of natural disasters: earthquakes \citep{earthquakes2020} and fires \citep{fires2020}. These distributions are represented on the sphere $\mathbb{S}^2$. Their complexity and multimodality make them suitable test cases for assessing MCF's usefulness in real world scenarios. Figure~\ref{fig:geo} shows the model's results. The density learned by MCF generally captures the modes and patterns in the data and can serve as a modelling tool which can be subject to further refinement by domain experts. As baselines, we trained NCPS and NMODEs, the same models we trained on spherical densities in section~\ref{sec:synthetic}, but could not achieve satisfactory results. We include them for completeness along with different spherical projections in Appendix~\ref{app:geo}. 
    
    
\subsection{Qualitative experiments: Lorenz attractor}
    
    Next, we model a distribution residing on a topologically non-trivial manifold. We illustrate our model's ability to preserve the ``global" manifold structure and learn the probability density of the Lorenz system by training on points along sampled trajectories. The stable manifold of the system's trajectories is a genus 2 manifold embedded in $\mathbb{R}^3$. For the classical parameter values, the Lorenz attractor admits a Sinai-Ruelle-Bowen (SRB) measure with support over the surface of the system \citep{tucker2002rigorous}. Informally, we can say that initial values ``diffuse'' over this surface. \looseness=-1
    
    To create an i.i.d.\@ dataset we generated 100 trajectories using the classical parameter values for the system, then uniformly sampled positions $\vx(t) \in \R^3$ for $t \in [0, 1000]$ along these. The procedure for the creation of the data set matches that of \citet{brehmer2020flows}. Our model consists of flows comprising five layers of RQ coupling transformations and models the manifold using two coordinate charts. More details on architectures and hyperparameter settings can be found in Appendix~\ref{app:lorenz}.
    
    Fig.~\ref{fig:lorenz} shows the manifold and probability distribution learned by our model and $\M$-flow. Parameterizing the manifold with multiple charts allows MCF to preserve the global topological structure of the manifold and to learn the probability distribution on it, even though the surface is self-intersecting. The single charted $\M$-flow struggles to accurately reconstruct the manifold, as it tries to cover the surface with a single coordinate chart, which implies the surface is homeomorphic to the plane. We do note however that $\M$-flow has captured the coarse-grained topological features of the surface, e.g. the reconstructed manifold is still genus 2 (i.e. contains two ``holes").\looseness=-1
    
    \begin{figure}[h]
        \includegraphics[width=\textwidth]{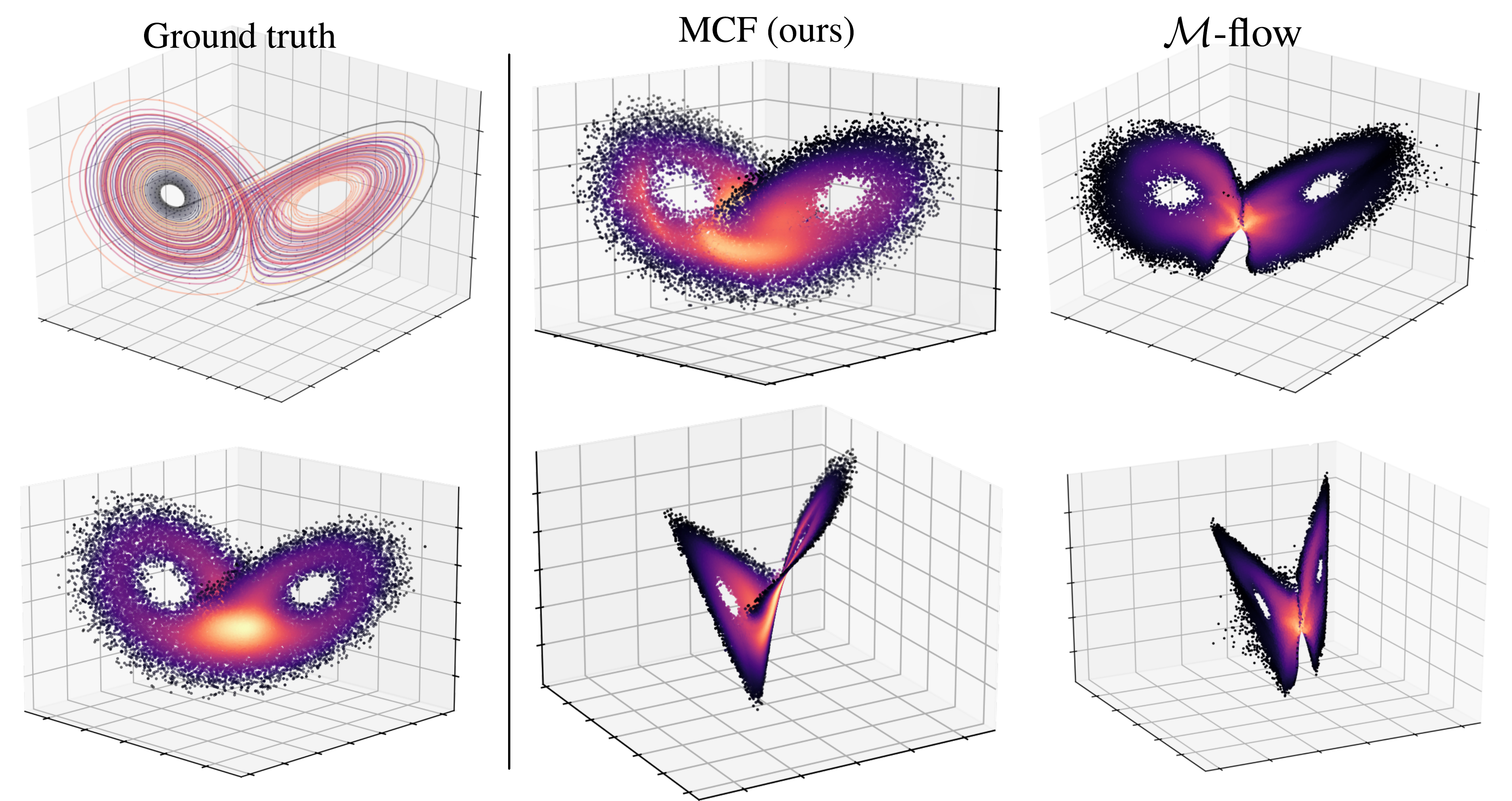}
        \caption{The manifold and probability distribution for the Lorenz attractor system. For the $\M$-flow model, the depicted manifold is learned, whereas for MCF the manifold shape is implicitly preserved through locally invertible models. Brighter color represents areas of higher estimated density. Ground truth shows sampled trajectories and the implicit surface formed by the system. }
        \label{fig:lorenz}
    \end{figure}
    
\subsection{Quantitative experiments: Real world particle physics data}
    Our quantitative experiment focuses on the task of inferring the parameters of a proton-proton collision process at the Large Hadron Collider (LHC). Raw data is usually in the order of millions, but following common practice among domain experts, we use a vector of 40 features to represent the data. The model of the process is based on a simulator which generates data $\vx \in \R^{40}$ given parameters $\boldsymbol{\theta} \in \R^3$ according to an implicit probability distribution $p(\vx|\boldsymbol{\theta})$. From domain experts we know that the data resides in a 14-dimensional manifold embedded in $\R^{40}$. Given the observations $\vx$ and parameters $\theta$, our task is to infer the posterior distribution over the parameters $p(\boldsymbol{\theta}|\vx)$. Thus, we train our model as a conditional density estimator to learn the simulator likelihood function.  
    
    Baseline models include a Euclidean flow in the ambient space (RQ-Flow, \citet{durkan19}), the $\M$-flow model, as well as an $\M$-flow variant with an unrestricted encoder denoted by $\M_e$-flow, both of which were proposed by \citet{brehmer2020flows}. Furthermore, \citet{brehmer2020flows} introduced versions of the models trained with the SCANDAL method \citep{brehmer2020mining}, which improves inference performance. All baselines are composed of thirty-five RQ coupling layers, interspersed with invertible, LU-decomposed linear transformations. The RQ-Flow is trained with maximum likelihood, while the $\M$-flow models are trained in two phases, as in \citet{brehmer2020flows} corresponding to a manifold learning phase and a density estimation phase. Our model (MCF) comprises five coordinate charts and each coordinate map is composed of ten RQ coupling layers, interspersed with invertible, LU-decomposed linear transformations. For further details on architectures and hyperparameters, see Appendix~\ref{app:lhc}.
    
    \begin{table}[h]
        \centering
        \begin{tabular}{ccc}
        Model & sample closure $\downarrow$ & log posterior $\uparrow$ \\
        \toprule
        RQ-Flow \citep{durkan19} & \textbf{0.0019} $\pm$ 0.0001 & -3.94 $\pm$ 0.87  \\
        RQ-Flow (SCANDAL) & 0.0565 $\pm$ 0.0059 & -0.49 $\pm$ 0.09 \\
        $\M$-flow \citep{brehmer2020flows} & 0.0045 $\pm$ 0.0004 & -1.71 $\pm$ 0.30 \\
        $\M$-flow (SCANDAL) & 0.0045 $\pm$ 0.0004 & 0.11 $\pm$ 0.04 \\
        $\M_e$-flow \citep{brehmer2020flows} & 0.0046 $\pm$ 0.0002 & -1.44 $\pm$ 0.34 \\
        $\M_e$-flow (SCANDAL) & 0.0291 $\pm$ 0.0010 & 0.03 $\pm$ 0.09 \\
        \hline
        MCF [ours] & 0.0040 $\pm$ 0.001 & \textbf{0.55} $\pm$ 0.21 \\
        \end{tabular}
        \vspace{2mm}
        \caption{Quantitative results on the large hadron collider (LHC) data. \emph{Sample closure} measures sample quality (lower is better), log-posterior score $\log p(\boldsymbol{\theta}|\vx_{\text{obs}})$ measures quality of inference (higher is better). Each model is trained five times with independent initializations. The top and bottom values are removed and the mean is computed over the remaining runs. Best results are shown in bold. Baseline results by \citet{brehmer2020flows}.}
        \label{tab:lhc}
    \end{table}
    
    For model evaluation, first we investigate the generative capabilities of all models by evaluating a series of tests on model samples. These ``closure tests'' are a weighted sum of individual constraints encoding relationships (derived from domain knowledge) between dimensions in the observed vector, taking values in $[0, 1]$, where smaller values denote higher sample quality. Second, we measure the quality of the log posterior inference. Given a set of 20 observed samples $\vx_{\text{obs}} \sim p(\vx|\boldsymbol{\theta}^{\star})$, we evaluate model likelihood in an MCMC sampler to generate posterior samples $\theta \sim p(\boldsymbol{\theta}|\vx_{\text{obs}})$. To evaluate the posterior, we then use kernel density estimation with a Gaussian kernel. We evaluate all models for three different ground truth parameter points $\boldsymbol{\theta}^{\star}$. For more details on the experimental setting of the task, see \cite{brehmer2020flows}.
    
    Table~\ref{tab:lhc} summarizes results for the LHC data. While the RQ-Flow learns a good sampler for the observed data judging by the closure test score, it does not estimate the density well, as evidenced by the log posterior score. Maximum likelihood in the ambient space does not take manifold topology into account, rather it relies on models with enough capacity to map the data to a base distribution in the ambient space. To the extent the model manages to learn such a mapping, it will be an adequate data sampler but will concurrently lead to biased density estimates due to the mismatch in the volume measures. Conversely, models that learn (such as $\M$-flow) or preserve (such as MCF) the topological structure of the data, achieve more accurate density estimates. Using multiple charts, our method outperforms all baselines in log posterior scores. In terms of sample quality, our method yields marginally better results than the single-charted baselines, which could imply that the underlying manifold is homeomorphic to Euclidean space, meaning a single chart is enough to capture its topology, however using multiple charts is beneficial for density estimation.
    
    \begin{table}[t]
        \centering
        \begin{tabular}{c|ccc}
            \toprule
            Datasets & \multicolumn{3}{c}{Models} \\
            \hline
            & MCF (ours) & NMODE & NCPS \\
            \hline
            Wrapped normals $(\mathbb{S}^2)$ & \textbf{1.53} $\pm 0.64$ &  $54.28 \pm 4.01$ &$2.69 \pm 0.42$ \\
            Checkerboard $(\mathbb{S}^2)$ & \textbf{3.40} $\pm 0.22$ & $50.81 \pm 2.74$ & $8.13 \pm 0.82$ \\
            \hline
            &  &  &  \\
            & MCF (ours) & \multicolumn{2}{c}{$\M$-flow} \\
            \hline
            Lorenz attractor & \textbf{16.35} $\pm 0.28$ & \multicolumn{2}{c}{$35.84 \pm 0.91$} \\
            \hline
            &  &  &  \\
            & MCF (ours) & $\M$-flow & RQ-Flow \\
            \hline
            Large Hadron Collider & \textbf{80.5} $\pm 1.32$ & $96.61\pm1.93$ & $99.74\pm4.71$ \\
            \bottomrule
        \end{tabular}
        \vspace{2.5mm}
        \caption{Model wallclock time per dataset (in hours). Computed over 3 training runs.}
        \label{tab:runtimes}
    \end{table}
    
    \subsection{Running times} \label{sec:runtimes}
    Respecting the topology of the data manifold yields tangible benefits to runtimes. Our approach generally uses fewer flow layers and parameters than most baselines leading to consistently smaller convergence times than all other baselines. Table~\ref{tab:runtimes} shows model wallclock times. For the experiments on section~\ref{sec:synthetic} all models were trained on the CPU. For all other experiments all models were trained on a Titan X (Pascal) GPU. 

\section{Conclusion} \label{sec:conclusion}
    We have presented a flow-based framework for modelling data distributions on non-Euclidean manifolds. Recent works in this direction either encode the topology of the target manifold in the model's architecture, rely on operations that do not scale to high dimensions or can, in principle, only learn Euclidean manifolds. In contrast, our method can generalize to manifolds of higher dimensions and/or complex topology. Our approach can converge faster and to better optima compared to most baselines. Shorter convergence times are not surprising since our approach does not require a lot of capacity to learn subsets of the manifold with simpler topology. ODE-based models and models that exploit local geometry match or surpass the performance of our approach, since geometric operations respect manifold topology, therefore providing a strong inductive bias. However, these models are inherently at a disadvantage regarding computational cost and scalability, since they rely either on sequential solvers, not taking full advantage of parallelization or on approximations of local operators that become prohibitively expensive in higher dimensions. Against models with structural priors, our model converges faster and achieves better optima since it does not rely on classical projective maps (like the cylindrical projection used by \citet{rezende20a}) which do not preserve topology.
    
    \paragraph{Limitations.}
    Optimizing multi-charted manifold flow models is consistently harder than their Euclidean counterparts, making hyperparameter configuration an important consideration. We also assume that the data resides on a smooth manifold, which might not necessarily be true. Finally, quantitative comparisons with other models become harder as different chart parameterizations of manifolds result in different units for the estimated log-likelihood.

\newpage
\bibliography{references}

\begin{thebibliography}{38}
\providecommand{\natexlab}[1]{#1}
\providecommand{\url}[1]{\texttt{#1}}
\expandafter\ifx\csname urlstyle\endcsname\relax
  \providecommand{\doi}[1]{doi: #1}\else
  \providecommand{\doi}{doi: \begingroup \urlstyle{rm}\Url}\fi

\bibitem[Beitler et~al.(2021)Beitler, Sosnovik, and Smeulders]{beitler2021pie}
Jan~Jetze Beitler, Ivan Sosnovik, and Arnold Smeulders.
\newblock Pie: Pseudo-invertible encoder.
\newblock \emph{arXiv preprint arXiv:2111.00619}, 2021.

\bibitem[Boomsma et~al.(2008)Boomsma, Mardia, Taylor, Ferkinghoff-Borg, Krogh,
  and Hamelryck]{boomsma2008generative}
Wouter Boomsma, Kanti~V Mardia, Charles~C Taylor, Jesper Ferkinghoff-Borg,
  Anders Krogh, and Thomas Hamelryck.
\newblock A generative, probabilistic model of local protein structure.
\newblock \emph{Proceedings of the National Academy of Sciences}, 105\penalty0
  (26):\penalty0 8932--8937, 2008.

\bibitem[Bose et~al.(2020)Bose, Smofsky, Liao, Panangaden, and
  Hamilton]{bose20a}
Joey Bose, Ariella Smofsky, Renjie Liao, Prakash Panangaden, and Will Hamilton.
\newblock Latent variable modelling with hyperbolic normalizing flows.
\newblock In Hal~Daumé III and Aarti Singh, editors, \emph{Proceedings of the
  37th International Conference on Machine Learning}, volume 119 of
  \emph{Proceedings of Machine Learning Research}, pages 1045--1055, Virtual,
  13--18 Jul 2020. PMLR.
\newblock URL \url{http://proceedings.mlr.press/v119/bose20a.html}.

\bibitem[Brand(2002)]{DBLP:conf/nips/Brand02}
Matthew Brand.
\newblock Charting a manifold.
\newblock In Suzanna Becker, Sebastian Thrun, and Klaus Obermayer, editors,
  \emph{Advances in Neural Information Processing Systems 15 [Neural
  Information Processing Systems, {NIPS} 2002, December 9-14, 2002, Vancouver,
  British Columbia, Canada]}, pages 961--968. {MIT} Press, 2002.
\newblock URL
  \url{https://proceedings.neurips.cc/paper/2002/hash/8929c70f8d710e412d38da624b21c3c8-Abstract.html}.

\bibitem[Brehmer and Cranmer(2020)]{brehmer2020flows}
Johann Brehmer and Kyle Cranmer.
\newblock Flows for simultaneous manifold learning and density estimation.
\newblock \emph{arXiv preprint arXiv:2003.13913}, 2020.

\bibitem[Brehmer et~al.(2020)Brehmer, Louppe, Pavez, and
  Cranmer]{brehmer2020mining}
Johann Brehmer, Gilles Louppe, Juan Pavez, and Kyle Cranmer.
\newblock Mining gold from implicit models to improve likelihood-free
  inference.
\newblock \emph{Proceedings of the National Academy of Sciences}, 117\penalty0
  (10):\penalty0 5242--5249, 2020.

\bibitem[Cornish et~al.(2020)Cornish, Caterini, Deligiannidis, and
  Doucet]{cornish20a}
Rob Cornish, Anthony Caterini, George Deligiannidis, and Arnaud Doucet.
\newblock Relaxing bijectivity constraints with continuously indexed
  normalising flows.
\newblock In Hal~Daumé III and Aarti Singh, editors, \emph{Proceedings of the
  37th International Conference on Machine Learning}, volume 119 of
  \emph{Proceedings of Machine Learning Research}, pages 2133--2143, Virtual,
  13--18 Jul 2020. PMLR.
\newblock URL \url{http://proceedings.mlr.press/v119/cornish20a.html}.

\bibitem[Dinh et~al.(2019)Dinh, Sohl{-}Dickstein, Pascanu, and
  Larochelle]{DinhRADApproach2019}
Laurent Dinh, Jascha Sohl{-}Dickstein, Razvan Pascanu, and Hugo Larochelle.
\newblock A {RAD} approach to deep mixture models.
\newblock \emph{CoRR}, abs/1903.07714, 2019.
\newblock URL \url{http://arxiv.org/abs/1903.07714}.

\bibitem[Dupont et~al.(2019)Dupont, Doucet, and Teh]{NEURIPS2019_21be9a4b}
Emilien Dupont, Arnaud Doucet, and Yee~Whye Teh.
\newblock Augmented neural odes.
\newblock In H.~Wallach, H.~Larochelle, A.~Beygelzimer, F.~d\textquotesingle
  Alch\'{e}-Buc, E.~Fox, and R.~Garnett, editors, \emph{Advances in Neural
  Information Processing Systems}, volume~32, pages 3140--3150. Curran
  Associates, Inc., 2019.
\newblock URL
  \url{https://proceedings.neurips.cc/paper/2019/file/21be9a4bd4f81549a9d1d241981cec3c-Paper.pdf}.

\bibitem[Durkan et~al.(2019)Durkan, Bekasov, Murray, and
  Papamakarios]{durkan19}
Conor Durkan, Artur Bekasov, Iain Murray, and George Papamakarios.
\newblock Neural spline flows.
\newblock In H.~Wallach, H.~Larochelle, A.~Beygelzimer, F.~d\textquotesingle
  Alch\'{e}-Buc, E.~Fox, and R.~Garnett, editors, \emph{Advances in Neural
  Information Processing Systems}, volume~32, pages 7511--7522. Curran
  Associates, Inc., 2019.
\newblock URL
  \url{https://proceedings.neurips.cc/paper/2019/file/7ac71d433f282034e088473244df8c02-Paper.pdf}.

\bibitem[EOSDIS(2020)]{fires2020}
EOSDIS.
\newblock Active fire data.
\newblock
  \url{https://earthdata.nasa.gov/earth-observation-data/near-real-time/firms/active-fire-data},
  2020.
\newblock Land, Atmosphere Near real-time Capability for EOS (LANCE) system
  operated by NASA’s Earth Science Data and Information System (ESDIS).

\bibitem[Falorsi and Forr{\'e}(2020)]{falorsi2020neural}
Luca Falorsi and Patrick Forr{\'e}.
\newblock Neural ordinary differential equations on manifolds.
\newblock \emph{arXiv preprint arXiv:2006.06663}, 2020.

\bibitem[Fefferman et~al.(2016)Fefferman, Mitter, and
  Narayanan]{fefferman2016testing}
Charles Fefferman, Sanjoy Mitter, and Hariharan Narayanan.
\newblock Testing the manifold hypothesis.
\newblock \emph{Journal of the American Mathematical Society}, 29\penalty0
  (4):\penalty0 983--1049, 2016.

\bibitem[Gemici et~al.(2016)Gemici, Rezende, and
  Mohamed]{gemici2016normalizing}
Mevlana~C Gemici, Danilo Rezende, and Shakir Mohamed.
\newblock Normalizing flows on riemannian manifolds.
\newblock \emph{arXiv preprint arXiv:1611.02304}, 2016.

\bibitem[Grathwohl et~al.(2018)Grathwohl, Chen, Bettencourt, Sutskever, and
  Duvenaud]{grathwohl2018ffjord}
Will Grathwohl, Ricky~TQ Chen, Jesse Bettencourt, Ilya Sutskever, and David
  Duvenaud.
\newblock Ffjord: Free-form continuous dynamics for scalable reversible
  generative models.
\newblock \emph{arXiv preprint arXiv:1810.01367}, 2018.

\bibitem[Hamelryck et~al.(2006)Hamelryck, Kent, and
  Krogh]{hamelryck2006sampling}
Thomas Hamelryck, John~T Kent, and Anders Krogh.
\newblock Sampling realistic protein conformations using local structural bias.
\newblock \emph{PLoS Comput Biol}, 2\penalty0 (9):\penalty0 e131, 2006.

\bibitem[Hutchinson(1989)]{hutchinson1989stochastic}
Michael~F Hutchinson.
\newblock A stochastic estimator of the trace of the influence matrix for
  laplacian smoothing splines.
\newblock \emph{Communications in Statistics-Simulation and Computation},
  18\penalty0 (3):\penalty0 1059--1076, 1989.

\bibitem[Karpatne et~al.(2018)Karpatne, Ebert-Uphoff, Ravela, Babaie, and
  Kumar]{karpatne2018machine}
Anuj Karpatne, Imme Ebert-Uphoff, Sai Ravela, Hassan~Ali Babaie, and Vipin
  Kumar.
\newblock Machine learning for the geosciences: Challenges and opportunities.
\newblock \emph{IEEE Transactions on Knowledge and Data Engineering},
  31\penalty0 (8):\penalty0 1544--1554, 2018.

\bibitem[Kingma and Ba(2014)]{kingma2014adam}
Diederik~P Kingma and Jimmy Ba.
\newblock Adam: A method for stochastic optimization.
\newblock \emph{arXiv preprint arXiv:1412.6980}, 2014.

\bibitem[Lee(2013)]{lee2013smooth}
John~M Lee.
\newblock Smooth manifolds.
\newblock In \emph{Introduction to Smooth Manifolds}, pages 1--31. Springer,
  2013.

\bibitem[Loshchilov and Hutter(2017)]{loshchilov2017decoupled}
Ilya Loshchilov and Frank Hutter.
\newblock Decoupled weight decay regularization.
\newblock \emph{arXiv preprint arXiv:1711.05101}, 2017.

\bibitem[Lou et~al.(2020)Lou, Lim, Katsman, Huang, Jiang, Lim, and
  De~Sa]{lou2020neural}
Aaron Lou, Derek Lim, Isay Katsman, Leo Huang, Qingxuan Jiang, Ser-Nam Lim, and
  Christopher De~Sa.
\newblock Neural manifold ordinary differential equations.
\newblock \emph{arXiv preprint arXiv:2006.10254}, 2020.

\bibitem[Mathieu and Nickel(2020)]{mathieu2020riemannian}
Emile Mathieu and Maximilian Nickel.
\newblock Riemannian continuous normalizing flows.
\newblock \emph{arXiv preprint arXiv:2006.10605}, 2020.

\bibitem[M\"{u}ller et~al.(2019)M\"{u}ller, Mcwilliams, Rousselle, Gross, and
  Nov\'{a}k]{muller19}
Thomas M\"{u}ller, Brian Mcwilliams, Fabrice Rousselle, Markus Gross, and Jan
  Nov\'{a}k.
\newblock Neural importance sampling.
\newblock \emph{ACM Trans. Graph.}, 38\penalty0 (5), October 2019.
\newblock ISSN 0730-0301.
\newblock \doi{10.1145/3341156}.
\newblock URL \url{https://doi.org/10.1145/3341156}.

\bibitem[Nascimento et~al.(2014)Nascimento, Silva, Marques, and
  Lemos]{DBLP:journals/tip/NascimentoSML14}
Jacinto~C. Nascimento, Jorge~G. Silva, Jorge~S. Marques, and Jo{\~{a}}o~Miranda
  Lemos.
\newblock Manifold learning for object tracking with multiple nonlinear models.
\newblock \emph{{IEEE} Trans. Image Process.}, 23\penalty0 (4):\penalty0
  1593--1605, 2014.
\newblock \doi{10.1109/TIP.2014.2303652}.
\newblock URL \url{https://doi.org/10.1109/TIP.2014.2303652}.

\bibitem[NOAA(2020)]{earthquakes2020}
NOAA.
\newblock Ncei/wd5 global significant earthquake database.
\newblock \url{https://www.ngdc.noaa.gov/hazard/earthqk.shtml}, 2020.
\newblock National Geophysical Data Center / World Data Service (NGDC/WDS).

\bibitem[Papamakarios et~al.(2017)Papamakarios, Pavlakou, and
  Murray]{papamakarios2017masked}
George Papamakarios, Theo Pavlakou, and Iain Murray.
\newblock Masked autoregressive flow for density estimation.
\newblock \emph{arXiv preprint arXiv:1705.07057}, 2017.

\bibitem[Papamakarios et~al.(2021)Papamakarios, Nalisnick, Rezende, Mohamed,
  and Lakshminarayanan]{DBLP:journals/jmlr/PapamakariosNRM21}
George Papamakarios, Eric~T. Nalisnick, Danilo~Jimenez Rezende, Shakir Mohamed,
  and Balaji Lakshminarayanan.
\newblock Normalizing flows for probabilistic modeling and inference.
\newblock \emph{J. Mach. Learn. Res.}, 22:\penalty0 57:1--57:64, 2021.
\newblock URL \url{http://jmlr.org/papers/v22/19-1028.html}.

\bibitem[Peel et~al.(2001)Peel, Whiten, and McLachlan]{peel2001fitting}
David Peel, William~J Whiten, and Geoffrey~J McLachlan.
\newblock Fitting mixtures of kent distributions to aid in joint set
  identification.
\newblock \emph{Journal of the American Statistical Association}, 96\penalty0
  (453):\penalty0 56--63, 2001.

\bibitem[Pitelis et~al.(2013)Pitelis, Russell, and
  Agapito]{DBLP:conf/cvpr/PitelisRA13}
Nikolaos Pitelis, Chris Russell, and Lourdes Agapito.
\newblock Learning a manifold as an atlas.
\newblock In \emph{2013 {IEEE} Conference on Computer Vision and Pattern
  Recognition, Portland, OR, USA, June 23-28, 2013}, pages 1642--1649. {IEEE}
  Computer Society, 2013.
\newblock \doi{10.1109/CVPR.2013.215}.
\newblock URL \url{https://doi.org/10.1109/CVPR.2013.215}.

\bibitem[Rezende and Mohamed(2015)]{pmlr-v37-rezende15}
Danilo Rezende and Shakir Mohamed.
\newblock Variational inference with normalizing flows.
\newblock In Francis Bach and David Blei, editors, \emph{Proceedings of the
  32nd International Conference on Machine Learning}, volume~37 of
  \emph{Proceedings of Machine Learning Research}, pages 1530--1538, Lille,
  France, 07--09 Jul 2015. PMLR.
\newblock URL \url{http://proceedings.mlr.press/v37/rezende15.html}.

\bibitem[Rezende et~al.(2020)Rezende, Papamakarios, Racaniere, Albergo, Kanwar,
  Shanahan, and Cranmer]{rezende20a}
Danilo~Jimenez Rezende, George Papamakarios, Sebastien Racaniere, Michael
  Albergo, Gurtej Kanwar, Phiala Shanahan, and Kyle Cranmer.
\newblock Normalizing flows on tori and spheres.
\newblock In Hal~Daumé III and Aarti Singh, editors, \emph{Proceedings of the
  37th International Conference on Machine Learning}, volume 119 of
  \emph{Proceedings of Machine Learning Research}, pages 8083--8092, Virtual,
  13--18 Jul 2020. PMLR.
\newblock URL \url{http://proceedings.mlr.press/v119/rezende20a.html}.

\bibitem[Roy et~al.(2007)Roy, Kemp, Mansinghka, and Tenenbaum]{roy2007learning}
Daniel~M Roy, Charles Kemp, Vikash~K Mansinghka, and Joshua~B Tenenbaum.
\newblock Learning annotated hierarchies from relational data.
\newblock In \emph{Advances in neural information processing systems}, pages
  1185--1192, 2007.

\bibitem[Rozen et~al.(2021)Rozen, Grover, Nickel, and Lipman]{rozen2021moser}
Noam Rozen, Aditya Grover, Maximilian Nickel, and Yaron Lipman.
\newblock Moser flow: Divergence-based generative modeling on manifolds.
\newblock \emph{Advances in Neural Information Processing Systems}, 34, 2021.

\bibitem[Schonsheck et~al.(2019)Schonsheck, Chen, and Lai]{schonsheck2019chart}
Stefan Schonsheck, Jie Chen, and Rongjie Lai.
\newblock Chart auto-encoders for manifold structured data.
\newblock \emph{arXiv preprint arXiv:1912.10094}, 2019.

\bibitem[Steyvers and Tenenbaum(2005)]{steyvers2005large}
Mark Steyvers and Joshua~B Tenenbaum.
\newblock The large-scale structure of semantic networks: Statistical analyses
  and a model of semantic growth.
\newblock \emph{Cognitive science}, 29\penalty0 (1):\penalty0 41--78, 2005.

\bibitem[Strichartz(2003)]{strichartz2003guide}
Robert~S Strichartz.
\newblock \emph{A guide to distribution theory and Fourier transforms}.
\newblock World Scientific Publishing Company, 2003.

\bibitem[Tucker(2002)]{tucker2002rigorous}
Warwick Tucker.
\newblock A rigorous ode solver and smale’s 14th problem.
\newblock \emph{Foundations of Computational Mathematics}, 2\penalty0
  (1):\penalty0 53--117, 2002.

\end{thebibliography}

\newpage
\appendix
\section*{Appendix}

\section{Proof of the lower bound on the data manifold log likelihood} \label{app:proof}
We will denote with $\M$ the data manifold of dimension $d$ embedded in some higher dimensional Euclidean ambient space $\R^D$. An open cover $\mathcal{U}$ of $\M$ consists of $K$ local coordinate charts $(U_i, \phi_i)_{i=1}^{K}$ with $U_i \subset \M$ and $\phi_i: U_i \rightarrow V_i \subset \R^d$. A probability density function $p_\M: \M \rightarrow \R$ can be constructed with a \emph{smooth partition of unity subordinate to} $\mathcal{U}$. That is, an indexed family $\{f_i\}_{i=1}^K$ of smooth functions with $\text{supp} f_i \subset U_i$, where for a neighborhood around any data point $\vx \in \M$, only a finite subset of $\{f_i\}$ is non-zero and $\sum_{i=1}^K f_i(x) = 1$. For our particular case, we take $f_i = w_i p_{U_i}$ with $w_i \in [0, 1]$ and construct $p_\M$ as a weighted sum of smooth density functions defined locally in each coordinate patch $U_i$, i.e. $p_\M(\vx) = \sum_{i=1}^K w_i p_{U_i} = \sum_i^K w_i p_{V_i}(\vu) \det |G_i(\vu)|^{-\frac{1}{2}}$, with $p_{V_i}$ the base distribution in Euclidean subset $V_i$, $\vu = \phi_i(\vx)$ and $G_i(\vu) = J_{\phi_{i^{-1}}}^T J_{\phi_i^{-1}}$.

\begin{proposition}
    The log-likelihood $\log p_\M(\vx)$ is bounded from below by $\mathcal{L} = \log \left[ C \cdot \sum_i^K w_i p_{V_i}(\vu) Tr(J_{\phi_i^{-1}}^{T}(\vu) J_{\phi_i^{-1}}(\vu))^{-\frac{d}{2}} \right]$ with $C = d^{d/2}$
\end{proposition}


    \paragraph{Proof}
    For a local coordinate chart $(U, \phi)$, with $U \subset \M$ with $\M \subset \R^D$ and $\phi: \M \rightarrow \R^d$, we denote the log-likelihood in neighborhood $U$ by $\log p_{U}(\vx)$. Furthermore, denoting the singular values of matrix $J_{\phi^{-1}}^T(\vu)$ by $s_i$, we will use Jensen's inequality to first lower-bound the probability density in a local neighborhood $U$:
    
    \begin{align}
        \frac{1}{2} \sum_{i=1}^{d} \log s_i^2 = \frac{d}{2} \sum_{i=1}^{d} \frac{1}{d} \log s_i^2 &\leq \frac{d}{2} \log \left ( \sum_{i=1}^d \frac{s_i^2}{d} \right ) = \frac{d}{2} \log \left ( \sum_{i=1}^d s_i^2 \right ) - \frac{d \log (d)}{2} \\
        -\frac{1}{2} \sum_{i=1}^{d} \log s_i^2 &\geq - \frac{d}{2} \log \left ( \sum_{i=1}^d s_i^2 \right ) + \frac{d \log (d)}{2} \\
        \log p_V(\vu) -\frac{1}{2} \sum_{i=1}^{d} \log s_i^2 &\geq \log p_V(\vu) - \frac{d}{2} \log \left ( \sum_{i=1}^d s_i^2 \right ) + \frac{d \log (d)}{2} \\
        \log p_V(\vu) - \frac{1}{2} \log[J_{\phi^{-1}}^T(\vu) J_{\phi^{-1}}(\vu)] &\geq \log p_V(\vu) - \frac{d}{2} \log \left ( \sum_{i=1}^d s_i^2 \right ) + \frac{d \log (d)}{2} \\
        \log p_{U}(\vx) &\geq \log p_V(\vu) - \frac{d}{2} \log \left ( \sum_{i=1}^d s_i^2 \right ) + \frac{d \log (d)}{2} \label{eq:lower-bound-singular}
    \end{align}
    
    Now for a matrix $A$, we have: 
    \begin{align}
        Tr(A^{T}A) = Tr(U^{T}\Sigma^{T} V V^{T} \Sigma U) = Tr(\Sigma^{T} \Sigma UU^{T}) = Tr(\Sigma^{T} \Sigma) = \sum_{i=1}^d s_i^{2}
    \end{align}
    
    with U, V orthogonal matrices and $\Sigma$ a diagonal matrix containing the singular values of $A$.
    
    Thus, eq.~\ref{eq:lower-bound-singular} becomes:
    \begin{align}
        \log p_U(\vx) &\geq \log p_V(\vu) - \frac{d}{2} \log Tr(J_{\phi^{-1}}^{T}(\vu) J_{\phi^{-1}}(\vu)) + \frac{d \log (d)}{2} = \mathcal{L}_U
    \end{align}
    
    Thus, we have introduced a lower bound to the probability density in neighborhood $U \in \M$. Because $\log(\cdot)$ is a monotonic function and for all $i$ we have $w_i \in [0, 1]$, the direction of the inequality in eq.~\ref{eq:lower-bound-singular} is preserved for all $K$ neighborhoods in our open cover of $\M$, so our lower bound holds for the complete data log likelihood:
    
    \begin{align}
        \log p(\vx) &= \log \sum_i^K w_i p_{U_i}(\vx) \\ 
        &= \log \sum_i^K w_i p_{V_i}(\vu) \det |G_i(\vu)|^{-\frac{1}{2}} \\
        &\geq \log \left[ C \cdot \sum_i^K w_i p_{V_i}(\vu) Tr\left( J_{\phi_i^{-1}}^{T}(\vu) J_{\phi_i^{-1}}(\vu) \right)^{-\frac{d}{2}} \right] = \mathcal{L}
    \end{align}
    
    with $C = d^{d/2}$. Using Hutchinson's estimator we can compute the trace efficiently. With $J_{\phi^{-1}} \in \R^{D \times d}$ and $p(\boldsymbol{\epsilon})=\mathcal{N}(0, I_D)$:
    \begin{align}
        \mathcal{L}_U &\approx \log p_{V_i}(\vu) - \frac{d}{2} \log \mathbb{E}_{p(\boldsymbol{\epsilon})}\left[ \boldsymbol{\epsilon}^{T} J_{\phi^{-1}}^{T}(\vu) J_{\phi^{-1}}(\vu) \boldsymbol{\epsilon} \right] + \frac{d \log (d)}{2} \\
        &= \log p_{V_i}(\vu) - \frac{d}{2} \log \mathbb{E}_{p(\boldsymbol{\epsilon})}\left[ ||J_{\phi^{-1}}^{T} \boldsymbol{\epsilon}||_2^2 \right] + \frac{d \log (d)}{2}
    \end{align}

\section{Details on synthetic 2D experiments} \label{app:synthetic}

\paragraph{Datasets} For all target densities in Figure~\ref{fig:synthetic-sphere} we generated 50000 points for the train set and 10000 points for the validation set. For details on generating the datasets, see \cite{lou2020neural}.

\paragraph{Architectures} Table~\ref{tab:mcf-architecture-synthetic} shows the architecture details for MCF. All flow layers are implemented as rational quadratic coupling flows. In all cases, our base distributions $p_{V_i}$ are standard normals over the Euclidean spaces $V_i$. The distribution of the orthogonal directions to the manifold $p_{V_i^{\perp}}$ is a standard normal with $\sigma^2 = 0.01$. 

Baseline implementations are provided by \citet{lou2020neural} in \url{https://github.com/CUAI/Neural-Manifold-Ordinary-Differential-Equations}.

\begin{table}[h]
    \centering
    \begin{tabular}{c|cc}
        Hyperparameters & \multicolumn{2}{c}{Datasets} \\
        \hline
         & Checkerboard ($\mathbb{S}^2$) & Four wrapped Normals ($\mathbb{S}^2$) \\
        \hline
        Charts & 4 & 2 \\
        Chart flow layers & 3 & 2 \\
        Chart bins & 5 & 1 \\
        Spline range & [-3, 3] & [-4, 4] \\
        Linear transform & LU & LU \\
        ResNet layers (\& units) & 2 (64) & 2 (16)\\
        Activation & ReLU & ReLU \\
    \end{tabular}
    \vspace{2mm}
    \caption{Architecture details for MCF on all synthetic datasets.}
    \label{tab:mcf-architecture-synthetic}
\end{table}

\paragraph{Training} We train MCF using maximum likelihood. For the \emph{spherical checkerboard} dataset we train for 300 epochs and for the \emph{four wrapped normals} we train for 250 epochs. For both datasets we used a batch size of 256 with a learning rate of $3 \cdot 10^{-4}$.

To train NCPS we used a learning rate of $10^{-3}$ and a batch size of 200. For the \emph{spherical checkerboard} dataset we train for 10000 epochs, while for the \emph{four wrapped normals} dataset we train for 5000 epochs. 

For NMODE, on \emph{four wrapped normals} we used a batch size of 200 and a learning rate of $10^{-2}$, training for 600 epochs. For the \emph{spherical checkerboard} we used a batch size of 200 and a learning rate of $10^{-2}$, training for 700 epochs. 

All models are trained with the Adam optimizer \citep{kingma2014adam}. In general, we chose baseline hyperparameters such that we can have the fastest possible convergence without sacrificing training stability. Please note however that this is a different training setting to the one used for NMODE and the other baselines by \citet{lou2020neural}, as they generated a random batch of points on the manifold for every iteration, whereas in our case we generate a fixed amount of training points and iterate on those. We think that while this is a much harder training scenario, it's also a more realistic one.

\section{Details on real world 2D experiments} \label{app:geo}

\begin{table}[h]
    \centering
    \begin{tabular}{c|cc}
        Hyperparameters & \multicolumn{2}{c}{Datasets} \\
        \hline
         & \emph{Fires} & \emph{Earthquakes} \\
        \hline
        Charts & 2 & 2 \\
        Chart flow layers & 5 & 4 \\
        Chart bins & 10 & 16 \\
        Spline range & [-6, 6] & [-6, 6] \\
        Linear transform & LU & LU \\
        ResNet layers (\& units) & 2 (100) & 2 (64)\\
        Activation & CELU & CELU \\
    \end{tabular}
    \vspace{2mm}
    \caption{Architecture details for MCF on all synthetic datasets.}
    \label{tab:mcf-architecture-geo}
\end{table}

\subsection{Experimental details}

\paragraph{MCF} For architectural details please see table~\ref{tab:mcf-architecture-geo}. Our base distribution $p_{V_i}$ is a standard normal in the Euclidean spaces $V_i$. The distribution of the orthogonal directions to the manifold $p_{V_i^{\perp}}$ is a standard normal with $\sigma^2 = 0.01$.

\paragraph{Baselines} The architectures of both the NMODE and NCPS baselines are the same as in the synthetic datasets case.

\paragraph{Datasets} The \emph{fires} dataset consists of 66444 data points, while the \emph{earthquakes} dataset consists of 5883 data points at the time of writing. We shuffle both datasets and keep 80\% for the training sets and 20\% for the validation sets.  

\paragraph{Training on \emph{earthquakes}} We train MCF using maximum likelihood for 3000 epochs using the Adam optimizer with a batch size of 128 and a learning rate of $3 \cdot 10^{-4}$. Throughout training we annealed the learning rate using a cosine annealing schedule.

We train NCPS with the Adam optimizer for 10000 epochs with a learning rate of $10^{-3}$ and a batch size of 200.

We train NMODE with the Adam optimizer for 10000 epochs with a learning rate $3 \cdot 10^{-4}$ and a batch size of 500.

\paragraph{Training on \emph{fires}} We train MCF using maximum likelihood for 1000 epochs using the Adam optimizer with a batch size of 256 and a learning rate $10^{-4}$. Throughout training we annealed the learning rate using a cosine annealing schedule and clipped the gradient norm to 8.

We train NCPS with the Adam optimizer for 3000 epochs with a learning rate of $10^{-3}$ and a batch size of 200.

We train NMODE with the Adam optimizer for 600 epochs with a learning rate $3 \cdot 10^{-4}$ and a batch size of 500.

\begin{figure}[h]
    \centering
    \begin{tabular}{cc}
        Earthquakes & Fires \\
        \includegraphics[width=0.45\linewidth]{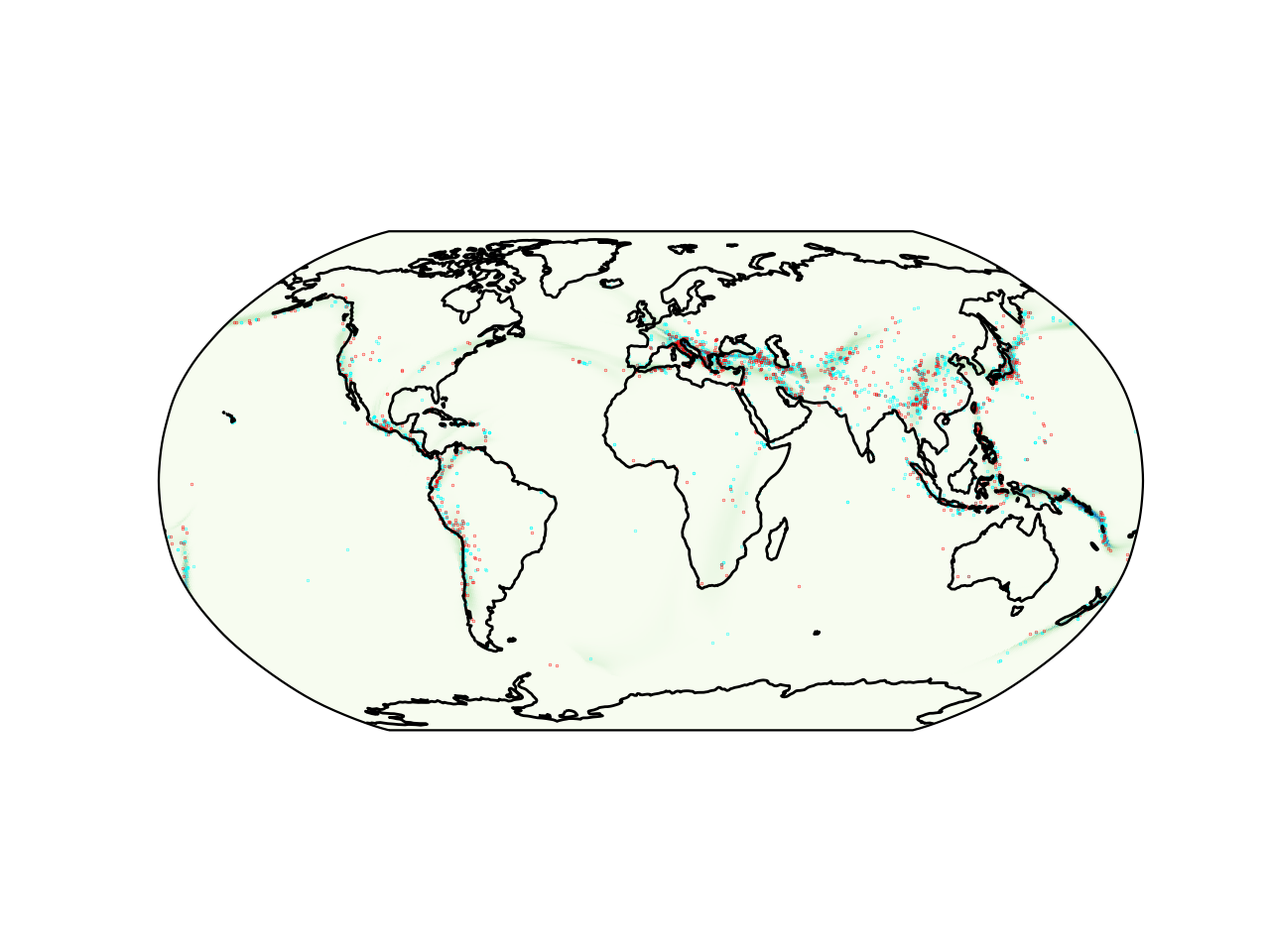} & \includegraphics[width=0.45\linewidth]{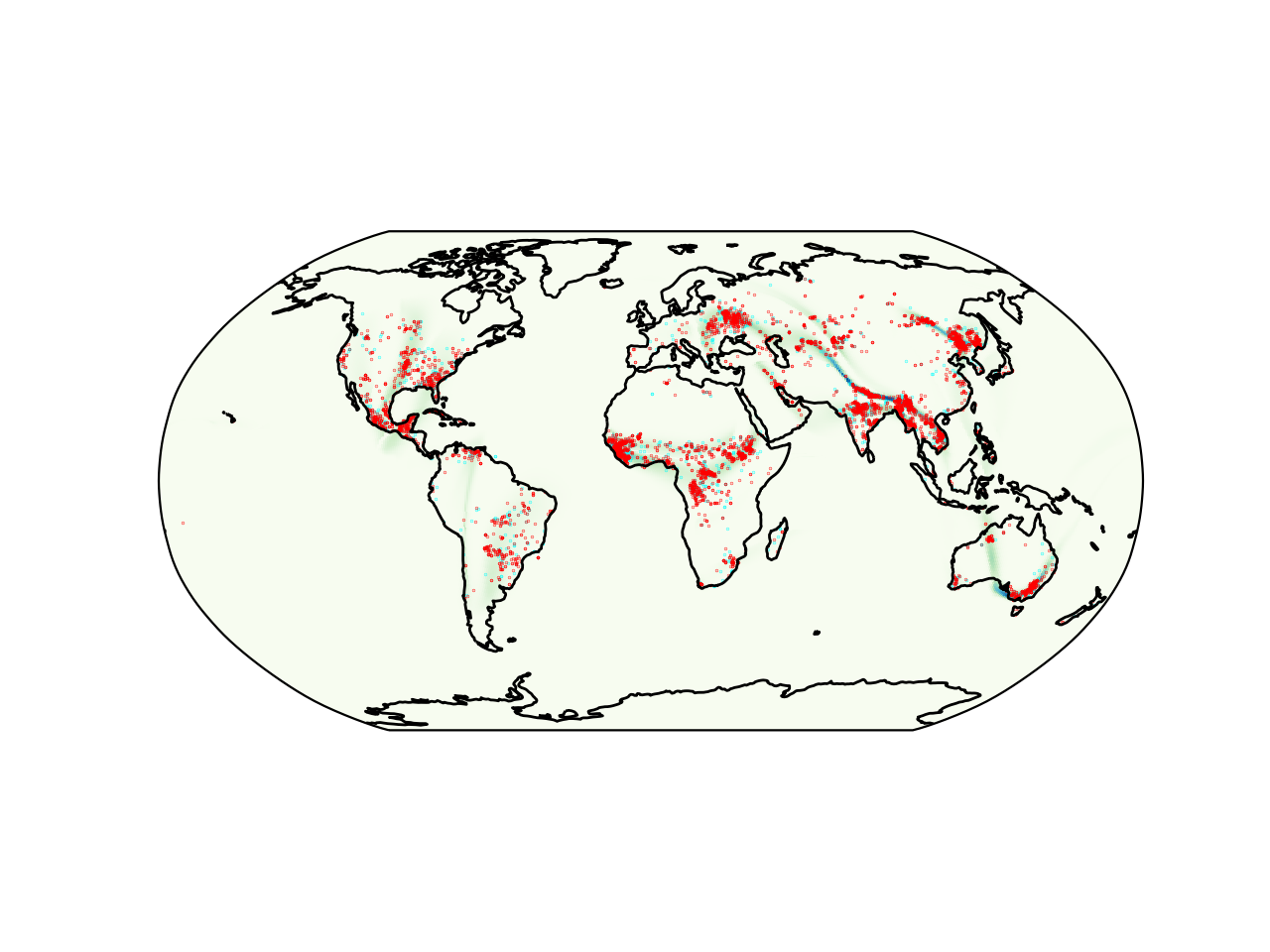} \\
        \includegraphics[width=0.45\linewidth]{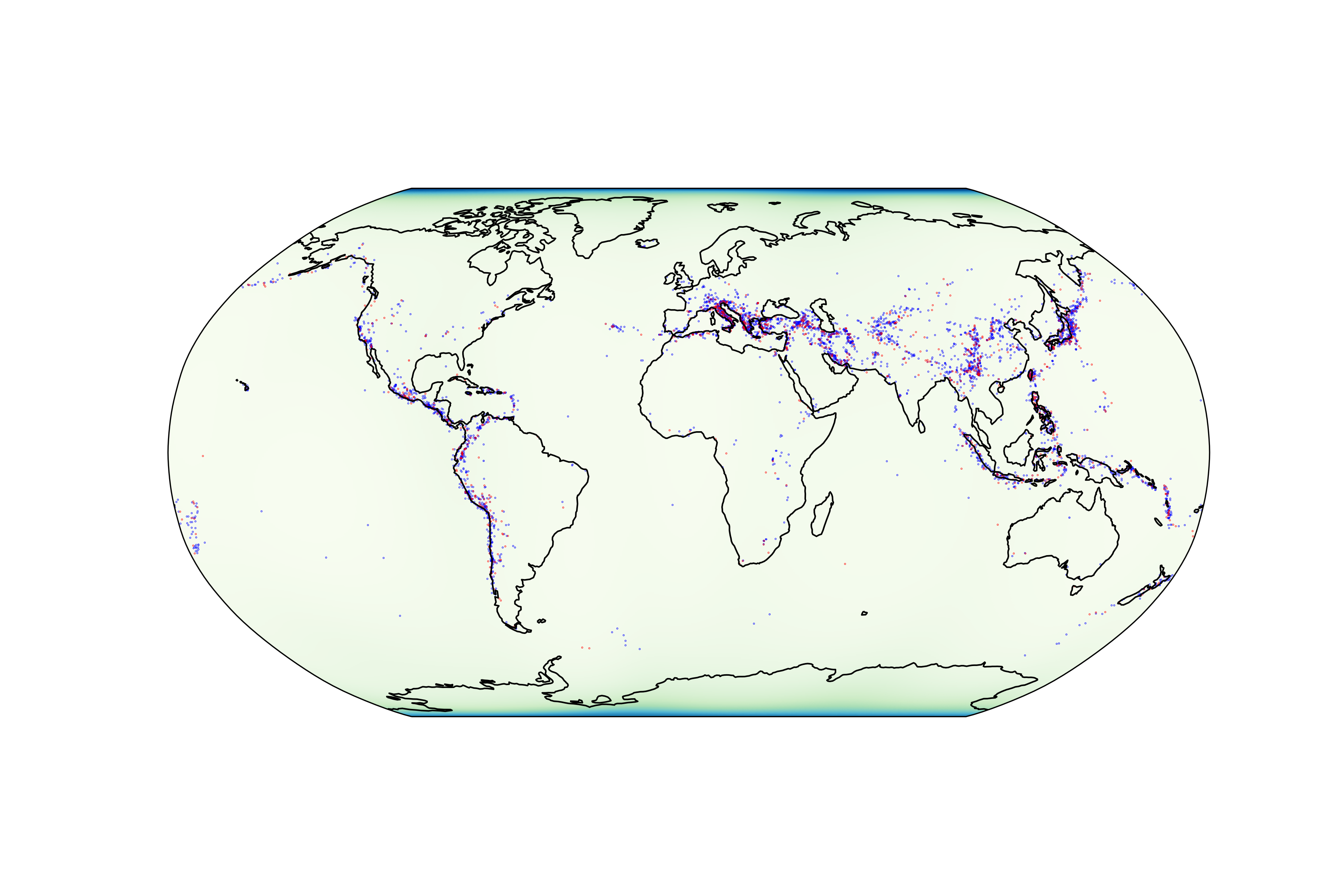} & \includegraphics[width=0.45\linewidth]{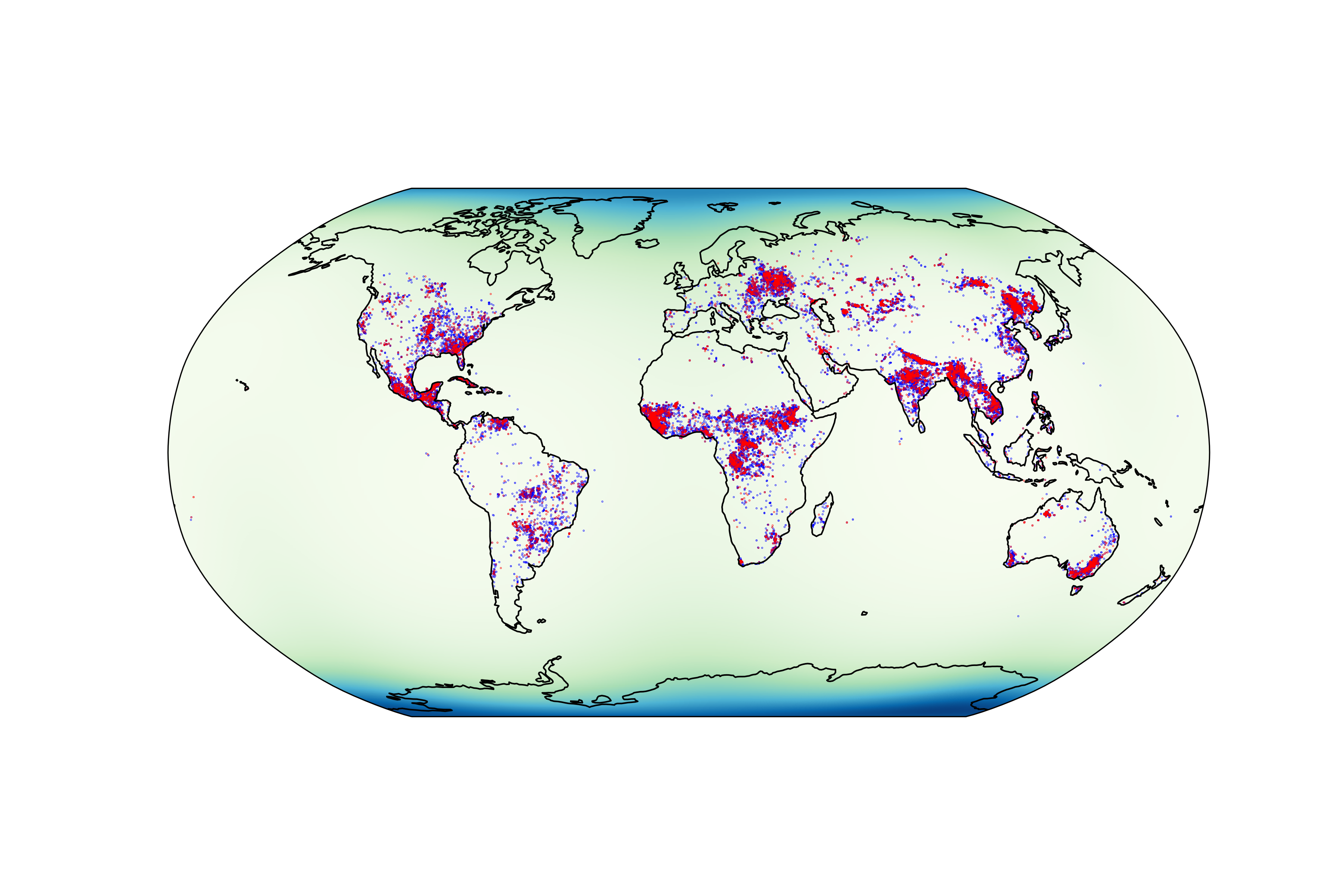} \\
        \includegraphics[width=0.45\linewidth]{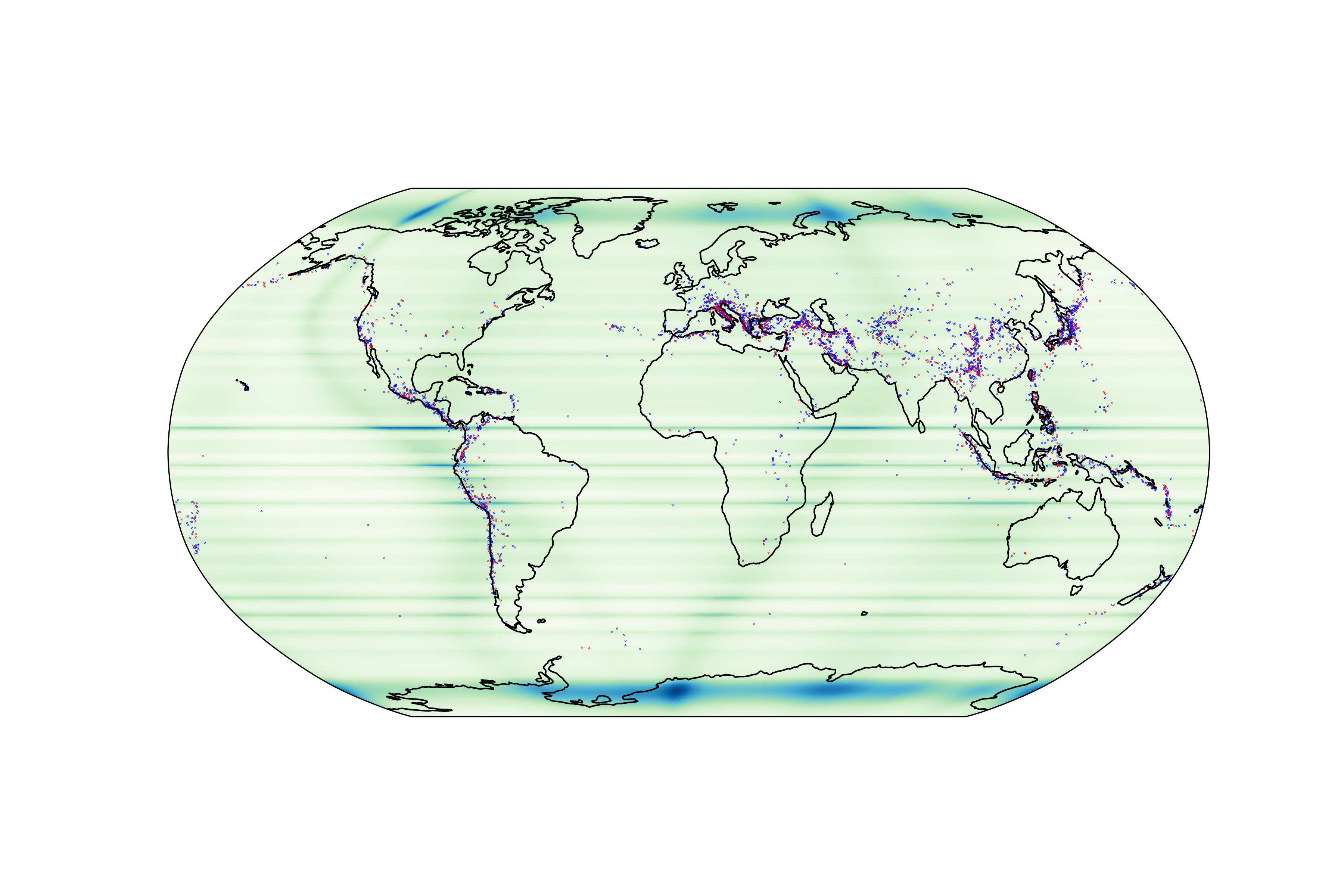} & \includegraphics[width=0.45\linewidth]{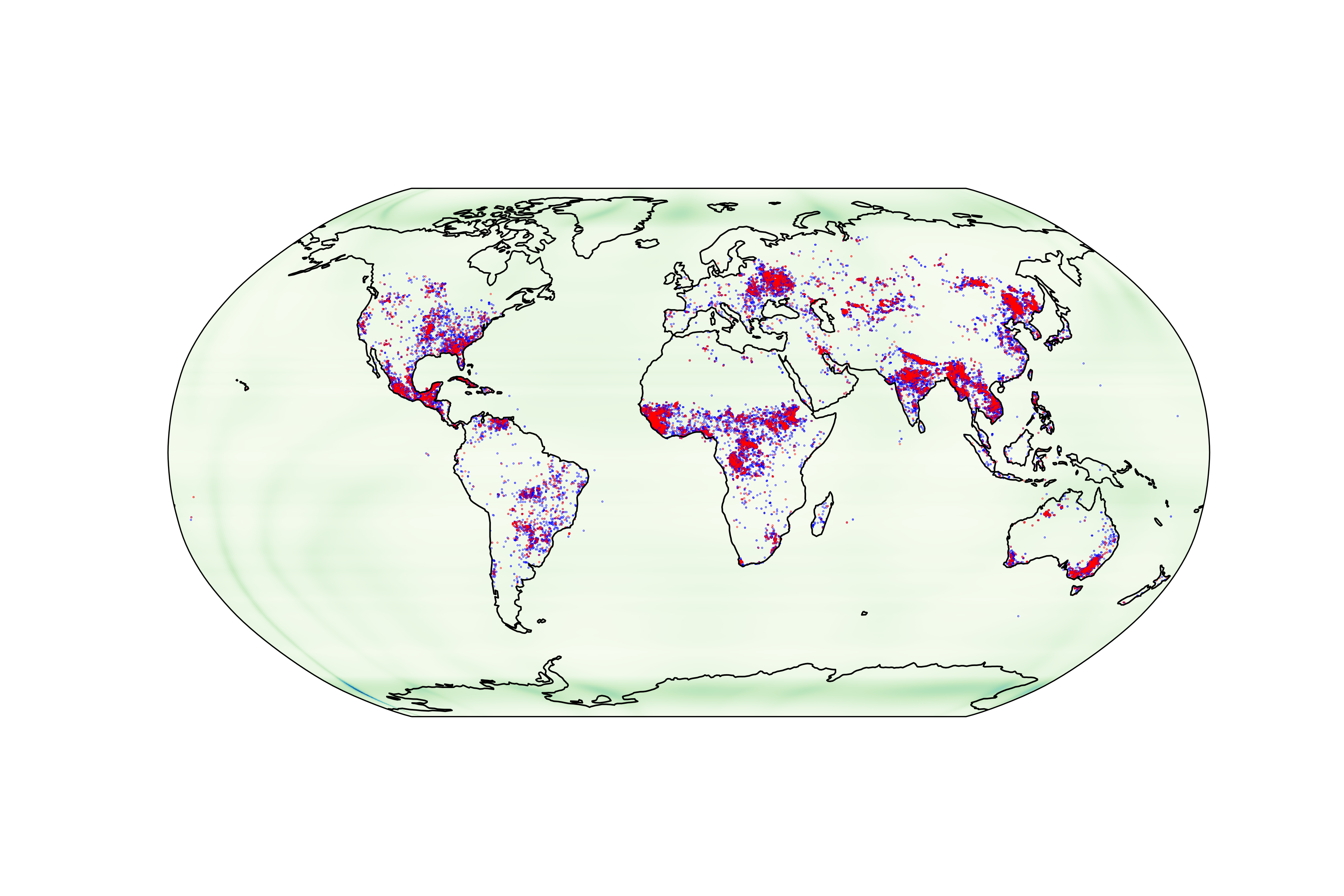}
    \end{tabular}
    \caption{Density estimation results on the earthquakes and fires data. Robinson projection. Top row: \textit{MCF} (ours), middle row: \textit{NMODE}, bottom row: \textit{NCPS}}
    \label{fig:disasters-extra-robinson}
\end{figure}

For all baselines in both datasets we decay the learning rate every 1/3d of the total epochs with a scaling factor of $0.1$. We found that training was difficult for all models. The hardest model to train was NMODE even though results for both baselines are generally unsatisfactory (see figure~\ref{fig:disasters-extra-robinson}). Given this, in our choice of hyperparameters we attempted to strike a balance between fast convergence and stable gradient updates. Furthermore, we checkpoint all models to retain the best performing parameter configuration according to validation results.

\section{Details on the Lorenz experiment} \label{app:lorenz}

\subsection{Architecture}
\paragraph{MCF} Our model uses two coordinate charts to parameterize the manifold. The chart models $\phi$ comprise five flow layers. These are implemented as rational quadratic coupling layers, interspersed with random feature permutations. We use five bins in the range $[-3, 3]$. Each coupling transform is parameterized by a residual network with 1 residual block containing 2 hidden layers per block. Each hidden layer consists of 32 ReLU units. Our base distribution $p_{V_i}$ is a standard normal over the Euclidean spaces $V_i$. The distribution of the orthogonal directions to the manifold $p_{V_i^{\perp}}$ is a standard normal with $\sigma^2 = 0.01$.

\paragraph{$\M$-flow} For $\M$-flow we reproduced the reference architecture given by \citet{brehmer2020flows}. Both the chart model and the base model comprise 5 rational quadratic coupling layers, interspersed with random feature permutations. We use 5 bins for both maps in the range [-3, 3]. Each coupling transform is parameterized by a residual network with 2 residual blocks and 2 hidden layers per block. Each hidden layer consists of 100 ReLU units.

\subsection{Training}

\paragraph{MCF} We trained the model on a dataset of $10^6$ samples using maximum likelihood training for 1000 epochs. The AdamW optimizer \citep{loshchilov2017decoupled} was used with a learning rate of $10^{-4}$. We use a batch size of 10000.

\paragraph{$\M$-flow} We trained the model on a dataset of $10^6$ samples with split manifold learning and maximum likelihood training phases, assigning 50 epochs to each phase (100 in total). The AdamW optimizer was used with a learning rate of $3 \cdot 10^{-4}$, cosine annealing and weight decay of $10^{-4}$. We use a batch size of 100. 

\section{Details on the Large Hadron Collider experiment} \label{app:lhc}

For details on dataset generation, as well as an explanation on the closure tests we refer the interested reader to \cite{brehmer2020flows}. The dataset itself can be found in \url{https://drive.google.com/drive/folders/13x8lEO8--L8-ORoN_QTUbSC_fRBAdRPT}.

\paragraph{MCF} Our model uses five coordinate charts to parameterize the manifold. The chart models $\phi$ comprise ten flow layers. These are implemented as rational quadratic coupling layers, interspersed with LU-decomposed invertible linear transformations. We use 11 bins in the range [-10, 10]. Each coupling transform is parameterized by a residual network with two residual blocks of two hidden layers per block. Each hidden layer consists of 100 ReLU units. Our base distributions $p_{V_i}$ are standard normals over the Euclidean spaces $V_i$. The distribution of the orthogonal directions to the manifold $p_{V_i^{\perp}}$ is a standard normal with $\sigma^2 = 0.01$.

\paragraph{Baselines} To estimate baseline runtimes we run both RQ-flow and $\M$-flow but we note that baseline results are taken from the paper itself. Both baselines are composed of 35 rational quadratic coupling layers, interspersed with LU-decomposed invertible linear transformations. For $\M$-flow, the chart model $\phi$ uses 20 layers and the base model $h$ uses 15 layers. Each coupling transform is parameterized by a residual network with two residual blocks of two hidden layers per block. Each hidden layer consists of 100 ReLU units. All runtime estimations are based on the implementation provided by \citet{brehmer2020flows}, which can be found in \url{https://github.com/johannbrehmer/manifold-flow}.

\paragraph{Training} We trained our model using maximum likelihood on the same dataset as \cite{brehmer2020flows}, using $10^6$ samples. We used the AdamW optimizer with a learning rate of $3 \cdot 10^{-4}$, a batch size of 256, cosine annealing and a weight decay of $10^{-5}$ and trained the model for 50 epochs. 

\paragraph{Evaluation} Our evaluation procedure is identical to \cite{brehmer2020flows}. In brief, we generate 3 different datasets using 3 different parameter points $\theta_1 = (0, 0), \theta_2 = (0.5, 0)$ and $\theta_3 = (-1, -1)$. Each dataset has 15 i.i.d.\@ samples. For each model and each observed dataset, we generate four MCMC chains of length 750 each, with a Gaussian proposal distribution with mean step size 0.15 and a burn in of 100 steps. Then we obtain kernel density estimates of the log-posterior for each of the 3 parameter points and report the average value in table~\ref{tab:lhc}. Like \citet{brehmer2020flows} we train 5 instances of our model with independent initializations, remove the top and bottom value and report the mean over the remaining runs.

\end{document}